\documentclass{article}

\usepackage[main, final]{neurips_2026}

\usepackage[utf8]{inputenc} 
\usepackage[T1]{fontenc}    
\usepackage{hyperref}       
\usepackage{url}            
\usepackage{booktabs}       
\usepackage{amsfonts}       
\usepackage{nicefrac}       
\usepackage{microtype}      
\usepackage{xcolor}         
\usepackage{xspace}
\usepackage{graphicx}       
\usepackage{pifont}         
\usepackage[nointegrals]{wasysym}  
\usepackage{caption}        
\usepackage{subcaption}     
\usepackage{array}          
\usepackage{multirow}       
\usepackage{makecell}       
\usepackage{longtable}

\newcommand{\sys}{PithTrain\xspace}
\newcommand{\sysbench}{ATE-Bench\xspace}
\title{\sys: A Compact and Agent-Native \\ MoE Training System}

\author{
  \textbf{Ruihang Lai}\textsuperscript{1}\thanks{Equal contribution} \quad
  \textbf{Hao Kang}\textsuperscript{1}$^*$ \quad
  \textbf{Haozhan Tang}\textsuperscript{1} \quad
  \textbf{Akaash R. Parthasarathy}\textsuperscript{1} \quad \\
  \textbf{Zichun Yu}\textsuperscript{1} \quad
  \textbf{Junru Shao}\textsuperscript{3} \quad
  \textbf{Todd C. Mowry}\textsuperscript{1} \quad
  \textbf{Chenyan Xiong}\textsuperscript{1,2}\thanks{Equal advising} \quad
  \textbf{Tianqi Chen}\textsuperscript{1,3}$^\dagger$ \quad \\
  \textsuperscript{1}Carnegie Mellon University \quad
  \textsuperscript{2}Xlue \quad
  \textsuperscript{3}NVIDIA \quad
}

\begin{document}

\newcommand{\cmark}{\textcolor[HTML]{2E7D32}{\ding{51}}}
\newcommand{\pmark}{\textcolor[HTML]{F9A825}{\LEFTcircle}}
\newcommand{\xmark}{\textcolor[HTML]{C62828}{\ding{55}}}

\newcommand{\megatron}{Megatron-LM\xspace}
\newcommand{\deepspeed}{DeepSpeed\xspace}
\newcommand{\torchtitan}{TorchTitan\xspace}

\newcommand{\vspacebeforecap}{\vspace{-1em}}
\newcommand{\vspaceaftercap}{\vspace{-1em}}

\newcommand{\MyPara}[1]{\vspace{.0em}\noindent\textbf{#1}}

\newcommand{\squishlist}{
   \begin{list}{$\bullet$}
    { \setlength{\itemsep}{1pt}      \setlength{\parsep}{3pt}
      \setlength{\topsep}{3pt}       \setlength{\partopsep}{0pt}
      \setlength{\leftmargin}{1em} \setlength{\labelwidth}{1em}
      \setlength{\labelsep}{0.5em} } }

\newcommand{\squishlisttwo}{
   \begin{list}{$\bullet$}
    { \setlength{\itemsep}{0pt}    \setlength{\parsep}{0pt}
      \setlength{\topsep}{0pt}     \setlength{\partopsep}{0pt}
      \setlength{\leftmargin}{2em} \setlength{\labelwidth}{1.5em}
      \setlength{\labelsep}{0.5em} } }

\newcommand{\squishend}{
    \end{list}  }

\maketitle
\setcounter{footnote}{0}

\begin{abstract}
Mixture-of-Experts (MoE) has become the dominant architecture for
frontier language models.
To meet this demand, production frameworks have built optimized
MoE training stacks over years of engineering effort.
Yet evolving these stacks for new architectures and system
optimizations remains expensive.
With the rise of AI coding agents, they could automate parts of
training-framework development and accelerate this evolution.
But applying them to these existing frameworks carries hidden
costs, invisible to today's throughput-only evaluations.
We name this missing dimension agent-task efficiency (ATE): the
cost of using coding agents to understand, operate, and extend a
framework.
Grounded in four agent-native design principles, we build \sys,
a compact, agent-native MoE training framework
{supporting pretraining, supervised fine-tuning, and reinforcement learning.}
We further introduce \sysbench, covering real-world
training-framework tasks.
{Our evaluation shows \sys achieves throughput competitive with production
frameworks and higher agent-task efficiency on \sysbench.
On new-feature tasks, \sys uses up to 62\% fewer Agent Turns than
Megatron-LM, and up to 44\% and 64\% less Active GPU Time than
Megatron-LM and TorchTitan, respectively.}

\vspace{1ex}
\par\noindent{\textbf{GitHub repo:} \url{https://github.com/mlc-ai/pith-train}}
\end{abstract}

\section{Introduction}
\label{sec:introduction}

Modern AI systems are increasingly powered by Mixture-of-Experts
(MoE) language models such as DeepSeek-V3, Qwen3, Kimi-K2, and
GPT-OSS~\cite{deepseekai2025deepseekv3technicalreport,yang2025qwen3technicalreport,openai2025gptoss120bgptoss20bmodel,kimiteam2026kimik2openagentic},
whose training depends on systems that scale across distributed
clusters. Production frameworks~\cite{shoeybi2020megatronlmtrainingmultibillionparameter,narayanan2021efficientlargescalelanguagemodel,deepspeed,rajbhandari2020zeromemoryoptimizationstraining} have built end-to-end MoE
training stacks over years of engineering effort, pairing layered
Python designs with extra compiled extensions to deliver broad
model coverage, peak throughput, and multi-platform support
needed for diverse training workloads.
However, evolving these stacks for new model architectures and
system optimizations demands in-depth expertise and substantial
engineering effort.

AI coding agents~\cite{claude-code,codex-cli,cursor,github-copilot}
have begun to automate parts of this work and could in principle
accelerate training-system development. Most current practice
applies agents to existing frameworks. But the design choices
that helped human engineers cost agents differently. Plugin
systems, registry-based indirection, and heavy compiled extensions
raise the cost of locating relevant code, tracing what runs at a
given call site, and verifying a change is complete. Today's
throughput-only evaluations leave this cost unmeasured, and
existing frameworks were not designed with it in mind.

This raises a design question: \emph{can we redesign an MoE
training framework that optimizes for agent-task efficiency?} We
define \emph{agent-task efficiency} (ATE) as the cost of using
coding agents to understand, operate, and extend a framework,
measured along dimensions such as session duration and output
tokens. We
answer this question with \sys, an end-to-end MoE training
framework designed agent-native from the start. \sys is built on
four design principles. First, we favor code compactness over
coverage: \sys focuses on a compact MoE training stack rather than
the broad model and feature coverage of production frameworks,
while remaining straightforward for agents to extend with new
features.
Second, we use minimal, Python-native components covering the key
layers of MoE training: operators, training engine, and
applications. Third, we use direct calls and avoid implicit
indirection in module composition, so that what runs at a given
call site can be identified by static reading. Finally, we ship
agent skills~\cite{anthropic-skills} for recurring
training-framework tasks.
{\sys supports pretraining and supervised fine-tuning (SFT),
and serves as a reinforcement learning (RL) backend in slime~\cite{slime}.}

Beyond building the framework, we systematically evaluate how
framework design affects agent-task efficiency on real
training-system tasks. Existing AI-coding
benchmarks~\cite{jimenez2024swebench,chen2021evaluatinglargelanguagemodels,austin2021programsynthesislargelanguage}
vary the agent on a fixed codebase to score agent capability,
focusing on general software-engineering tasks such as issue
resolution. We invert this with \sysbench, a comprehensive benchmark that varies
the framework on real-world training-framework tasks, holding
the agent fixed {to compare framework-level development cost,
including code structure, bundled guidance, and runtime behavior}. \autoref{fig:overview} summarizes \sys's
overall design.

\begin{figure}[!t]
  \centering
  \includegraphics[width=0.95\textwidth]{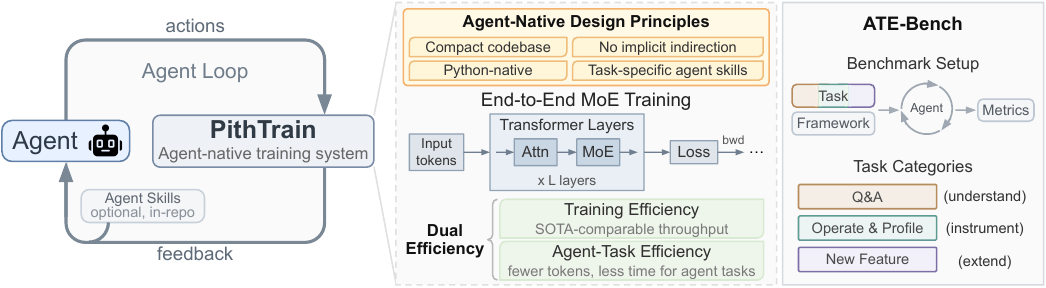}
  \caption{\sys overview. An agent issues actions and consumes
  feedback (left); built on four agent-native design principles,
  \sys delivers dual efficiency (middle); \sysbench evaluates
  agent-task efficiency across frameworks (right).}
  \label{fig:overview}
\end{figure}

This paper makes three contributions:

\squishlist
    \item \textbf{\sys, a compact, Python-native, end-to-end MoE
    training framework.} {Designed for coding agents, \sys achieves training throughput
    competitive with production frameworks.}
    \vspace{-2pt}
    \item \textbf{Four design principles for agent-native ML
    training frameworks}, guiding \sys's design: code compactness,
    Python-native components, no implicit indirection, and agent
    skills.
    \vspace{-2pt}
    \item \textbf{Agent-task efficiency, \sysbench, and an
    empirical study.} Agent-task efficiency as a metric beyond
    training throughput, instantiated in \sysbench, with an
    empirical study showing
    \sys's higher agent-task efficiency over production frameworks,
    plus a skills ablation and case study.
\squishend

\sys delivers \emph{dual efficiency}: strong training throughput
together with high agent-task efficiency. {It achieves throughput competitive with production frameworks across a range
of MoE models and training settings on NVIDIA H200 and B200 GPUs.
On \sysbench's new-feature tasks, a coding agent uses up to 62\% fewer
Agent Turns on \sys than on \megatron, and up to 44\% and 64\% less
Active GPU Time than on \megatron and \torchtitan, respectively.} A qualitative case
study illustrates how framework design
shapes agent behavior.

\section{Related Work}
\label{sec:related}

Mature production frameworks have driven large-scale MoE training.
\megatron~\cite{shoeybi2020megatronlmtrainingmultibillionparameter,narayanan2021efficientlargescalelanguagemodel}
established the pipeline-parallel Transformer recipe most
subsequent frameworks build on, and
\deepspeed~\cite{deepspeed,rajbhandari2020zeromemoryoptimizationstraining}
introduced ZeRO sharded-optimizer techniques. Both rely on
layered designs with plugin systems, registry-based indirection,
and compiled extension chains, delivering broad model coverage
and peak throughput on production hardware.
\torchtitan~\cite{liang2025torchtitanonestoppytorchnative} more
recently scaled PyTorch-native training to multi-thousand-GPU
clusters, sharing \sys's Python-native goal but without making
agent-native design a primary axis or matching production-framework
throughput on our configurations. \sys takes a different point on
this tradeoff curve with agent-native design as a first-class
goal, achieving higher agent-task efficiency along Agent Turns,
Output Tokens, etc., while {achieving training throughput competitive with}
production frameworks.

AI coding agents combine a tool-using language model with a small,
stable vocabulary of primitives (read, edit, shell, search).
Recent work designs better agent flows and harnesses for software
engineering on top of these
primitives~\cite{yao2023react,yang2024sweagentagentcomputerinterfacesenable,zhang2024autocoderoverautonomousprogramimprovement}.
\sys takes a complementary direction: rather than improving the
agent flow, we design the software framework so that existing
flows work better, lowering agent cost without changing the agent.

Existing agent benchmarks score capability on fixed codebases and
tasks: SWE-bench~\cite{jimenez2024swebench} on GitHub-issue
resolution, MLE-bench~\cite{chan2025mlebench} on Kaggle-style ML
engineering, and
HumanEval~\cite{chen2021evaluatinglargelanguagemodels} on
function-level code generation. Closer to ML systems engineering,
FlashInfer-Bench~\cite{xing2026flashinferbenchbuildingvirtuouscycle}
and KernelBench~\cite{ouyang2025kernelbench} target inference
operators and GPU kernels respectively. {Other benchmarks also
measure agent cost: GitTaskBench incorporates token cost into an
economic-benefit metric~\cite{ni2025gittaskbench}, SWE-QA reports token
usage for repository-level question answering~\cite{peng2025sweqa},
and public coding-agent leaderboards report token usage and cost
alongside performance~\cite{aa-coding-agents}. These evaluations vary
the agent.} \sysbench inverts this: holding the agent and task fixed,
we vary the training framework {to compare development cost on
ML training-system tasks, including Active GPU Time consumed during
testing and debugging}. This axis is complementary to existing
{agent-facing} benchmarks.

\section{The \sys System}
\label{sec:pithtrain}

\subsection{Agent-Native Design Principles}
\label{sec:pithtrain:principles}

This subsection introduces the four agent-native design principles
that guide \sys's system design, and contrasts how production
training frameworks align with each principle today.
\autoref{tab:principles-comparison} summarizes this comparison;
production frameworks adopt different subsets of the four
principles, reflecting different design priorities.
\begin{table}[t]
  \centering
  \small
  \caption{Adherence to agent-native design principles.
  \cmark{} satisfied; \pmark{} partial; \xmark{} not satisfied.
  Numbers are total framework LoC across Python, C++, and CUDA.\protect\footnotemark}
  \label{tab:principles-comparison}
  \begin{tabular}{lcccc}
    \toprule
    Framework & Compact & Python-native & No implicit indirection & Agent skills \\
    \midrule
    \megatron    & \xmark~149K  & \xmark & \xmark & \pmark \\
    \deepspeed   & \xmark~167K  & \xmark & \xmark & \xmark \\
    \torchtitan  & \pmark~38K   & \cmark & \pmark & \pmark \\
    \textbf{\sys} & \cmark~11K & \cmark & \cmark & \cmark \\
    \bottomrule
  \end{tabular}
\end{table}

\footnotetext{{\sysbench evaluated \sys \texttt{23db182}, \megatron \texttt{3bec9aa}, and \torchtitan \texttt{d84e83d}. \deepspeed: \texttt{44c51e3}.}}

\MyPara{Compact codebase.}
Production frameworks such as \megatron and \deepspeed offer broad
coverage of models, training features, and hardware platforms,
accumulated over years of engineering effort, with core codebases
exceeding {100K} lines. A larger codebase inflates the cost of
locating relevant code, tracking cross-file dependencies, and
verifying a change is complete. A compact codebase reduces this
cost; with frontier coding agents operating at context windows of
200K to 1M tokens, a sufficiently compact codebase can also fit in
a single context pass. We treat compactness as a constraint on
growth: \sys may grow over time, but {we keep the code that defines models and
coordinates training compact as kernel implementations grow.}

\MyPara{Python-native codebase.}
Python is the dominant language in modern ML.
A pure-Python framework lets an agent navigate the full framework
in a single language, surfaces readable Python tracebacks instead
of opaque native errors, and eliminates the compiled-extension
rebuild cycle. {These shorter feedback loops reduce the cost of
iterative debugging.} \megatron composes its core transformer layers from
out-of-tree TransformerEngine~\cite{transformer-engine} modules,
and \deepspeed bundles extensive in-tree extensions. These deliver
peak kernel performance and vendor-tuned numerics, but push an
agent across language boundaries and force a rebuild on change.

\begin{figure}[t]
  \centering
  \includegraphics[width=0.9\textwidth]{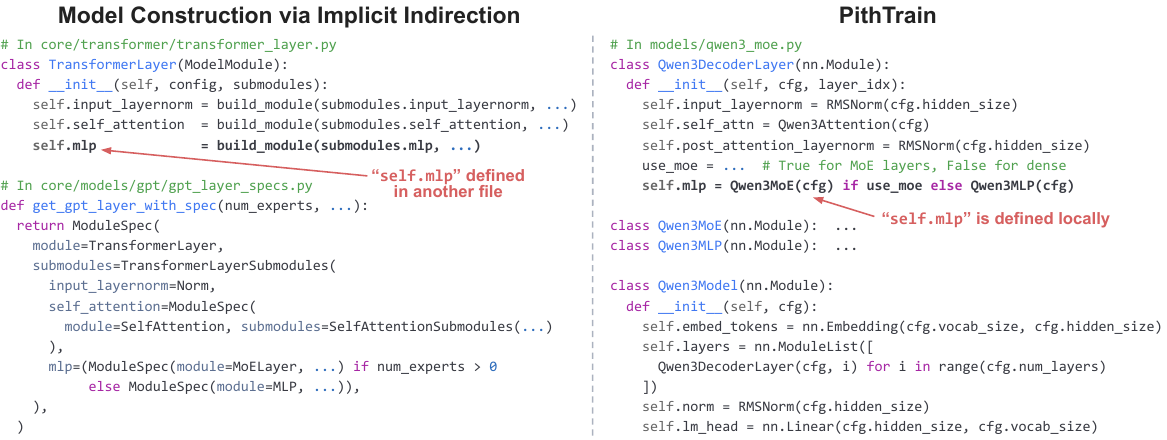}
\caption{Model construction patterns, illustrating the
  no-implicit-indirection principle. The implicit-indirection pattern
  resolves submodules through a runtime spec, supporting model
  variation from a shared layer skeleton; \sys instantiates
  layers directly, favoring local readability.}
  \label{fig:indirection-contrast}
\end{figure}

\MyPara{No implicit indirection.}
Production frameworks compose many model variants from a shared
layer skeleton via \emph{implicit indirection} (a stored
callable, plugin registry, or string-keyed resolution). This
pattern enables code reuse across models, while making what runs
at a given call site harder to identify by local reading.
\autoref{fig:indirection-contrast} shows an instance of model
construction in a production framework: \texttt{TransformerLayer}
resolves its submodules from a runtime spec in a separate file.
{An agent must resolve that spec before symbol search can reach
the implementation.} A flat code structure trades cross-model reuse for local
readability, reducing the effort an agent spends building an
end-to-end understanding.
{We balance this tradeoff through limited model-local duplication
and shared functions with explicit call targets.}

\MyPara{Task-specific agent skills.}
An \emph{agent skill}~\cite{anthropic-skills} is a procedural
playbook a coding agent loads on demand. Skills encode procedural knowledge an agent cannot recover from static reading alone, so it
runs a verified procedure. Agent
skills are a recent practice that existing training frameworks
have not yet adopted at scale: \megatron ships six skills for
CI/PR hygiene; \torchtitan ships one
developer-workflow skill plus four editor rules orienting the
agent to the codebase; \deepspeed ships two markdown files with
generic rules. None of these target recurring training-framework
tasks.

\subsection{System Architecture and Optimizations}
\label{sec:pithtrain:arch-and-opt}

\begin{figure}[t]
  \centering
  \includegraphics[width=0.75\textwidth]{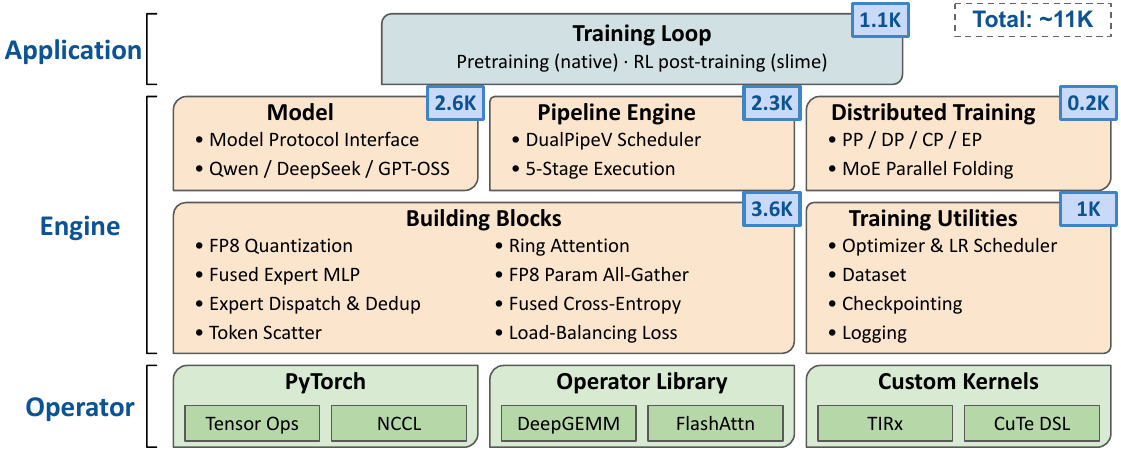}
  \caption[Architecture of \sys.]{{Architecture of \sys. Blue badges give per-component lines of code.}\protect\footnotemark}
  \label{fig:arch}
\end{figure}
\footnotetext{{Line counts are from \texttt{23db182}, before the adoption of TIRx and CuTe DSL.}}

\begin{table}[t]
  \centering
  \caption{{How \sys realizes each agent-native design principle.}}
  \label{tab:principle-map}
  {\small
  \begin{tabular}{@{}ll@{}}
    \toprule
    Principle & {Realization} \\
    \midrule
    Compact codebase & All boxes in \autoref{fig:arch}; line counts sum to $\sim$11K \\
    Python-native codebase & {Python throughout, including DSLs for kernels} \\
    No implicit indirection & {Flat structure; self-contained model files} \\
    {Task-specific agent skills} & Specialized skills under the project root \\
    \bottomrule
  \end{tabular}%
  }
\end{table}

The codebase is organized in three layers (application, engine, operator), as shown in \autoref{fig:arch}.
{We identified the components necessary for distributed MoE training.}
Where production
frameworks deliver broad out-of-the-box coverage of models,
features, and hardware, \sys narrows scope to keep the codebase
compact and reachable in a single context window.
\autoref{tab:principle-map} maps each principle to where it is
realized: we adopt a flat code structure with no plugin registries
or runtime specs, with each MoE model living in a self-contained
file under \texttt{models/} rather than being composed through a
shared layer skeleton. This favors local readability over cross-model
code reuse.

\sys supports standard MoE training\footnote{\sys also supports
dense (non-MoE) models, with MoE-specific machinery (e.g., expert
parallelism) skipped.} with pipeline parallelism (PP),
data parallelism (DP) via FSDP~\cite{pytorchfsdp}, context
parallelism (CP)~\cite{liu2024ringattention}, expert parallelism
(EP)~\cite{lepikhin2020gshardscalinggiantmodels}, FP8
training~\cite{micikevicius2022fp8formatsdeeplearning}, and DCP
checkpointing~\cite{torch-dcp}, on NVIDIA Hopper and Blackwell GPUs.
{Beyond pretraining, \sys serves as a training backend for RL in slime~\cite{slime}, supporting asynchronous
GRPO~\cite{shao2024deepseekmath}.}
Despite its compact, Python-native codebase, \sys aims for
training throughput competitive with production frameworks by
{adopting established MoE optimizations. We describe their integration
for transparency and reproducibility; the optimizations themselves are
not a separate contribution.}

\squishlist
\item \textbf{DualPipeV pipeline schedule.}
{\sys builds on DualPipe from
DeepSeek-V3~\cite{deepseekai2025deepseekv3technicalreport}, whose
open-source version provides only the pipeline scaffold; \sys adds the
compute--communication overlap. Each layer is split into five stages at
EP communication boundaries, and the schedule overlaps the forward of
one micro-batch with the backward of another to hide EP all-to-alls.}

\vspace{-2pt}

\item {\textbf{Agent-evolved GPU kernels.}
We use agents to evolve fast kernels for Hopper and Blackwell GPUs
with workflows such as TIRx Harness~\cite{tirx-harness} and KDA~\cite{kda}.
Each evolved kernel is validated against a reference implementation
before adoption, and kernels live in one file per architecture, so the
code an agent edits stays small and self-contained.}

\vspace{-2pt}

\item \textbf{Torch compile.}
\sys applies \texttt{torch.compile(fullgraph=True)} to all
transformer computation except the MoE forward and backward. This
strict mode rejects graph breaks at compile time rather than
silently degrading speedup. We exclude the MoE forward and
backward because per-expert input shapes are data-dependent under
expert parallelism.

\vspace{-2pt}

\item \textbf{Other optimizations.}
\sys also implements
{MoE parallel folding~\cite{liu2025moeparallelfolding} to decouple
attention and expert parallelism;}
EP
dispatch deduplication for lower all-to-all communication volume;
{FP8 parameter all-gather in FSDP to halve weight-gather traffic;}
and fused kernels {written in TIRx~\cite{tirx} and CuTe DSL~\cite{cute-dsl}}
for EP token scatter and FP8 quantization.
\squishend

\subsection{Agent Skills}
\label{sec:pithtrain:skills}

A skill encodes the procedure for one recurring training-framework
task. \sys ships a suite of skills covering several common ones
(Table~\ref{tab:skills}). Each skill is a self-contained folder with
a top-level \texttt{SKILL.md} playbook, optionally additional markdown
documents, and optionally helper scripts. Some skills are pure
markdown, like \texttt{add-new-model}; others bundle scripts that
offload deterministic work.
{Skills are composable: task-specific skills can invoke shared utility
skills to reuse common procedures.}
{These skills complement general improvements to agents and harnesses
with framework-specific procedures maintained alongside the code.}

\begin{figure}[t]
\centering
\begin{minipage}[c]{0.50\textwidth}
  \centering
  \captionof{table}{Representative skills in \sys.}
  \label{tab:skills}
  \vspace{2pt}
  {\footnotesize
  \renewcommand{\arraystretch}{1.4}%
  \begin{tabular}{@{}l@{\hspace{1.2em}}l@{}}
    \toprule
    {Agent Skill} & {Example Usage} \\
    \midrule
    \texttt{add-new-model} & {Port a new MoE architecture} \\
    \texttt{capture-nsys-profile} & {Inspect a training step} \\
    {\texttt{analyze-nsys-profile}} & {Find exposed communication} \\
    \texttt{validate-correctness} & {Verify a refactor keeps loss} \\
    {\texttt{validate-performance}} & {Benchmark an optimization} \\
    {\texttt{launch-with-slurm}} & {Run multi-node training} \\
    \bottomrule
  \end{tabular}%
  }
\end{minipage}%
\hfill
\begin{minipage}[c]{0.46\textwidth}
  \centering
  \includegraphics[width=0.99\textwidth]{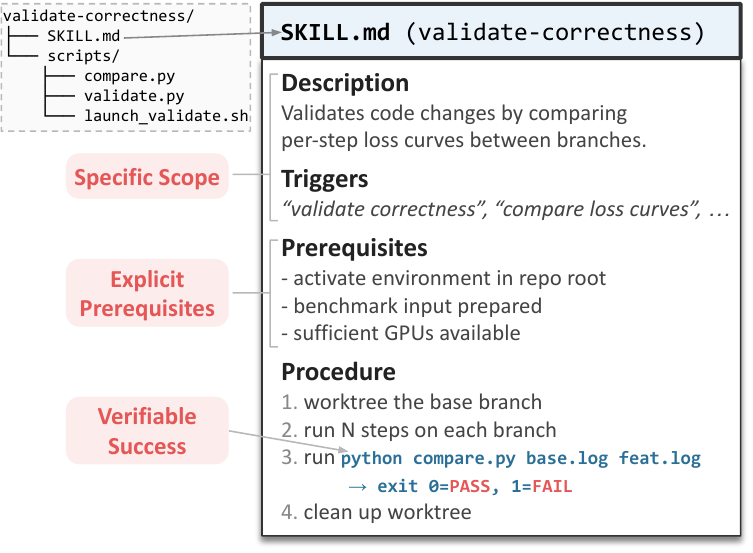}
  \caption{The \texttt{validate-correctness} skill in \sys. Pink labels mark properties.}
  \label{fig:skill-anatomy}
\end{minipage}
\vspace{-10pt}
\end{figure}

Each skill in \sys is designed around three properties:
\emph{specific scope}, \emph{explicit prerequisites}, and
\emph{verifiable success}. \autoref{fig:skill-anatomy} illustrates
these on \texttt{validate-correctness}. The description and triggers
together encode the skill's specific scope. The prerequisites section
enumerates environment, data, and configuration assumptions, so
missing state is caught before the skill begins to run. The procedure
ends in a script call that returns a reproducible PASS/FAIL verdict
rather than the agent's self-assessment. Skills designed around these
properties are not technically hard to author, and we expect other
training frameworks to ship comparable coverage as the practice
matures.
\section{\sysbench: A Benchmark for Agent-Task Efficiency}
\label{sec:benchmark}

Evaluating the agent-task efficiency of a training framework requires
varying the framework while holding the agent and task fixed. This is
the inverse of standard benchmarks for AI coding agents~\cite{jimenez2024swebench,chen2021evaluatinglargelanguagemodels,austin2021programsynthesislargelanguage},
which hold the codebase and task fixed and vary the agent to score
capability. We introduce \sysbench,
a benchmark with a fixed agent and curated task suite run across
frameworks {to compare their agent-task efficiency}. The suite spans the kinds of work
researchers typically perform on training frameworks, organized
around three recurring patterns: understanding the framework
without modifying it, operating it as a tool for instrumentation
and profiling, and extending it with new functionality. Tasks are distributed across three categories:
\squishlist
  \vspace{-3pt}
  \item Q\&A (12 tasks): answer questions whose answer is a
  property of the code, not a runtime measurement (e.g., ``how is
  the device mesh built?'').
  \vspace{-3pt}
  \item Operate and Profile (4 tasks): run, instrument, and
  profile the framework as a tool (e.g., capture an Nsight Systems
  profile and identify the most expensive CUDA kernels).
  \vspace{-3pt}
  \item New Feature (4 tasks): port a new model architecture into
  the framework end-to-end against a published reference
  implementation (e.g., Mixture of Block Attention
  (MoBA)~\cite{lu2025moba}).
  \vspace{-3pt}
\squishend

\autoref{fig:benchmark-harness} illustrates the agent loop for each
category, with agent involvement deepening from Q\&A (read-only)
to Operate and Profile (running, minor instrumentation) to New
Feature (substantial modification, test-driven iteration). Full
task descriptions and per-category correctness checks are in
Appendix~\ref{sec:appendix:tasks}. Using \sysbench, we evaluate
\sys and production frameworks, reporting five effort metrics:
session duration, active GPU time, agent turns, per-turn context,
and output tokens. Without a single-scalar metric for agent-task
efficiency, we report each dimension independently.

Q\&A questions are chosen to be valid across all three
frameworks, excluding framework-specific behaviors.
{New-feature tasks come from independent publications and reference
implementations; none was integrated into any evaluated framework at
study time.}
\sysbench does not cover tasks like cross-model propagation of
a shared change, where production frameworks' implicit
indirection may lower agent effort; we leave these as future work.

\begin{figure}[t]
  \centering
  \vspace{7pt}
  \includegraphics[width=0.85\textwidth]{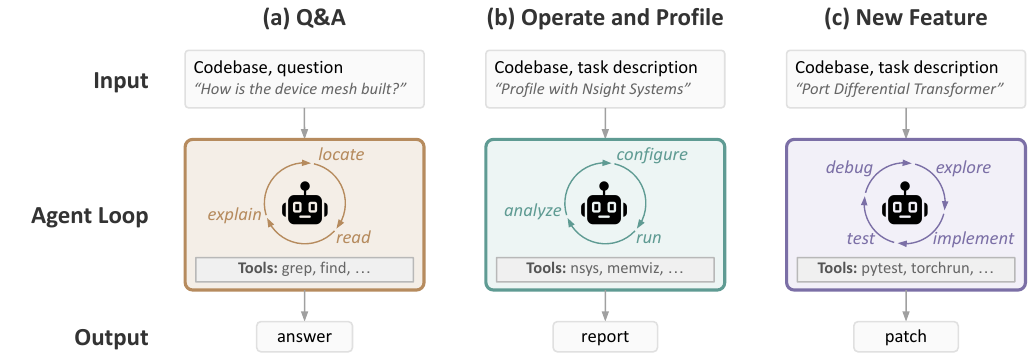}
  \caption{{Agent loop for each \sysbench task category, from
  read-only Q\&A to code-modifying New Feature tasks. Categories differ
  in loop steps, tools, and output.}}
  \vspace{-9pt}
  \label{fig:benchmark-harness}
\end{figure}

\section{Evaluation}
\label{sec:evaluation}

We evaluate \sys on both axes of dual efficiency: training
efficiency and agent-task efficiency. We validate training
correctness against \megatron on both pretraining loss curves and
downstream accuracy in
Appendix~\ref{sec:appendix:model-quality}.
{We also evaluate asynchronous GRPO on Qwen3-30B-A3B-Base,
where \sys achieves training reward and AIME 2024 accuracy comparable
to \megatron (Appendix~\ref{sec:appendix:rl}).}
This section is organized to answer the following questions:
\squishlist
  \vspace{-2pt}
  \item Does \sys deliver competitive training throughput against production frameworks?~(\S\ref{sec:eval:training})
  \vspace{-2pt}
  \item Does \sys offer higher agent-task efficiency than production frameworks?~(\S\ref{sec:eval:agent})
  \vspace{-2pt}
  \item How much do agent skills improve agent-task efficiency on \sys?~(\S\ref{sec:eval:skills-ablation})
  \vspace{-2pt}
  \item Where do the per-framework differences in agent cost come from
  on a single concrete task?~(\S\ref{sec:eval:case})
  \vspace{-2pt}
\squishend

\subsection{Training Efficiency}
\label{sec:eval:training}

\begin{table}[t]
  \centering
  \small
  \caption{Training throughput across frameworks.
  We report the aggregate tokens per second as the training throughput.
  {Parentheses give BF16 MFU~\cite{chowdhery2023palm} against a
  989.4~TFLOPS dense BF16 peak per H200; Appendix~\ref{sec:appendix:mfu}
  gives the FLOP count.}
  }
  \label{tab:training-throughput}
  \setlength{\tabcolsep}{3pt}
  \resizebox{\linewidth}{!}{%
  \begin{tabular}{lll|rrr}
    \toprule
    Model & Hardware & Parallelism / SeqLen / Precision & \megatron & \torchtitan & \sys \\
    \midrule
    {DeepSeek-V2-Lite} & {1$\times$8-H200} & {PP1,DP8,CP1,EP8 / 4K / BF16} & {204.3K (42.3\%)} & {171.0K (35.5\%)} & {193.5K (40.1\%)} \\
    {Qwen3-30B-A3B} & {2$\times$8-H200} & {PP2,DP8,CP1,EP8 / 4K / BF16} & {219.0K (31.9\%)} & {170.2K (24.8\%)} & {227.2K (33.1\%)} \\
    {Qwen3-30B-A3B} & {2$\times$8-H200} & {PP2,DP2,CP4,EP8 / 32K / BF16} & {100.9K (36.3\%)} & {89.6K (32.2\%)} & {107.2K (38.5\%)} \\
    {GPT-OSS-120B} & {4$\times$8-H200} & {PP4,DP8,CP1,EP8 / 4K / BF16} & {209.1K (21.6\%)} & {238.2K (24.6\%)} & {312.4K (32.3\%)} \\
    {Qwen3.5-35B-A3B} & {1$\times$8-B200} & {PP1,DP1,CP8,EP8 / 64K / FP8} & {86.6K} & {47.3K} & {118.3K} \\
    \bottomrule
  \end{tabular}}
\end{table}

We compare \sys~({\texttt{cdf6f0e}}), \megatron~\cite{narayanan2021efficientlargescalelanguagemodel}~({\texttt{685fb8b}})
and \torchtitan~\cite{liang2025torchtitanonestoppytorchnative}~({\texttt{45e33cd}}) on {four}
representative MoE models{:} {DeepSeek-V2-Lite~\cite{deepseekai2024deepseekv2strongeconomicalefficient},} Qwen3-30B-A3B~\cite{yang2025qwen3technicalreport},
{Qwen3.5-35B-A3B}, and {GPT-OSS-120B}~\cite{openai2025gptoss120bgptoss20bmodel}{,}
under matched parallelism (PP, DP, CP, EP), sequence length, and precision.
Configurations span single-node and multi-node deployments on
NVIDIA {H200} and B200 GPUs.
For \megatron, we follow NVIDIA's documented best practices\protect\footnote{\url{https://docs.nvidia.com/nemo/megatron-bridge/nightly/training/moe-optimization.html}}.
\deepspeed is excluded as it does not currently support PP combined with EP for MoE
training\protect\footnote{Verified at commit \texttt{44c51e3}: the
Megatron-DeepSpeed integration script states ``Currently we don't
support PP for MoE.''}.

To ensure the MoE router exhibits steady-state load-balanced
routing across frameworks and thus comparable throughput,
we load public model checkpoints rather than random initializations.
Each experiment runs 25 steps, and we report the median step
time over the last 10.
{We omit Model FLOPs Utilization (MFU)~\cite{chowdhery2023palm}
for the FP8 configuration because Tensor Core peak FLOPS differs
between BF16 and FP8, making the metric ambiguous for mixed-precision steps.} Training hyperparameters follow
Appendix~\ref{sec:appendix:model-quality}.

As \autoref{tab:training-throughput} shows, {\sys exceeds both baselines on four of five configurations
and stays within 5.3\% of \megatron on DeepSeek-V2-Lite.}
This {performance} comes from optimizations such
as DualPipeV's compute--communication overlap{, agent-evolved fast GPU kernels,}
and \texttt{torch.compile(fullgraph=True)}. These results demonstrate
that a compact, Python-native codebase can achieve competitive
training throughput.

\subsection{Agent-Task Efficiency}
\label{sec:eval:agent}

In this section, we evaluate agent-task efficiency across
frameworks.
We run \sysbench (\S\ref{sec:benchmark}) on \megatron, \torchtitan,
and \sys with Claude Code (Opus~4.7 at
\texttt{xhigh} effort\protect\footnote{\url{https://platform.claude.com/docs/en/build-with-claude/effort}})
as the fixed agent. {Operate-and-profile and new-feature tasks
run on a single node with eight NVIDIA H100 GPUs.}
Each task runs three times and we report medians;
{task descriptions, correctness criteria, and per-attempt values}
are described in {Appendices~\ref{sec:appendix:tasks} and
\ref{sec:appendix:claude}}. Opus~4.7 completed
{every task across all attempts and frameworks. We repeat the
suite with Codex and report its results and run-to-run variation in
Appendix~\ref{sec:appendix:codex} and~\ref{sec:appendix:dispersion}.}

\MyPara{Q\&A.}
Answering a question requires locating where a behavior lives in
the codebase. {We omit Session Duration} and Active GPU Time (no training runs). All 12
questions are answered correctly across attempts and frameworks
(grading details in
Appendix~\ref{sec:appendix:tasks:qa-correctness}).
Across the 12 questions, the agent uses up to 67\% fewer Agent Turns to reach the final answer on \sys than on \megatron.
{The lower Per-Turn Context is consistent with a smaller search
space (\autoref{tab:agent-efforts-codebaseqa}).}

\begin{table}[t]
  \centering
  \small
  \vspace{-7pt}
  \caption{Per-question agent effort on the Q\&A task suite
  (\S\ref{sec:benchmark}); Q1--Q12 are described in
  Appendix~\ref{sec:appendix-codebaseqa}. Each metric reports the
  median of three independent attempts. Lower is more efficient.
  Headers abbreviate \megatron, \torchtitan, and \sys.}
  \label{tab:agent-efforts-codebaseqa}
  \begin{tabular}{c|rrr|rrr|rrr}
    \toprule
       & \multicolumn{3}{c|}{Agent Turns}    & \multicolumn{3}{c|}{Per-Turn Context}            & \multicolumn{3}{c}{Output Tokens} \\
    \# & Megatron & Titan & Pith & Megatron & Titan & Pith & Megatron & Titan & Pith \\
    \midrule
    Q1  & 33         & 18          & \textbf{15} & 45.8K          & 44.3K          & \textbf{33.4K} & 9.0K           & 7.1K           & \textbf{4.1K}  \\
    Q2  & 54         & 28          & \textbf{18} & 41.2K          & 43.5K          & \textbf{35.9K} & 10.7K          & 7.8K           & \textbf{5.4K}  \\
    Q3  & 14         & 14          & \textbf{13} & \textbf{31.3K} & 36.1K          & 34.1K          & \textbf{3.4K}  & 3.5K           & 3.4K           \\
    Q4  & 36         & 26          & \textbf{16} & 41.3K          & 39.7K          & \textbf{31.1K} & 10.0K          & 6.9K           & \textbf{5.2K}  \\
    Q5  & 48         & \textbf{19} & 21          & 46.5K          & 41.7K          & \textbf{33.8K} & 10.7K          & 4.8K           & \textbf{4.6K}  \\
    Q6  & 12         & \textbf{4}  & 11          & 36.4K          & 29.9K          & \textbf{27.9K} & 4.5K           & \textbf{2.3K}  & 3.3K           \\
    Q7  & \textbf{9} & 12          & 14          & 31.5K          & \textbf{30.0K} & 31.5K          & \textbf{2.6K}  & 3.4K           & 4.2K           \\
    Q8  & 20         & 11          & \textbf{9}  & 30.8K          & 32.7K          & \textbf{28.7K} & 4.7K           & 3.5K           & \textbf{3.2K}  \\
    Q9  & 34         & \textbf{16} & \textbf{16} & 45.2K          & 38.6K          & \textbf{37.0K} & 7.6K           & 6.0K           & \textbf{5.1K}  \\
    Q10 & 19         & 17          & \textbf{11} & 37.8K          & 38.8K          & \textbf{29.4K} & 5.9K           & 5.1K           & \textbf{2.9K}  \\
    Q11 & 10         & 8           & \textbf{6}  & 31.3K          & 30.1K          & \textbf{25.6K} & 3.1K           & 2.5K           & \textbf{2.0K}  \\
    Q12 & 18         & \textbf{5}  & 10          & 37.1K          & 33.9K          & \textbf{31.1K} & 5.1K           & \textbf{2.2K}  & 2.6K           \\
    \bottomrule
  \end{tabular}
\end{table}

\MyPara{Operate and Profile.}
Across all tasks (\autoref{tab:agent-efforts-operate}), \sys's Agent Turns are up to 70\% lower than \megatron's and 57\% lower than \torchtitan's, and its Output Tokens are up to 78\% and 65\% lower respectively.
{These comparisons evaluate each framework's codebase together
with its in-repo documentation and skills.}
In addition, the agent invokes in-repo
skills (\S\ref{sec:pithtrain:skills}) on its own when applicable;
for example, the \emph{Report Heavy Kernels} task triggers
\texttt{capture-nsys-profile}.

\begin{table}[t]
  \centering
  \small
  \caption{Per-task agent effort across frameworks on
  operate-and-profile tasks (\S\ref{sec:benchmark}). Each metric
  reports the median of three independent attempts. Lower is more
  efficient. Session Duration and Active GPU Time are in minutes. Per-attempt metrics in {Appendix~\ref{sec:appendix:claude}}.}
  \label{tab:agent-efforts-operate}
  \setlength{\tabcolsep}{3pt}
  \begin{tabular}{@{}lr|rrrrr@{}}
    \toprule
    Task
      & Framework
      & \makecell[br]{Session\\Duration}
      & \makecell[br]{Active\\GPU Time}
      & \makecell[br]{Agent\\Turns}
      & \makecell[br]{Per-Turn\\Context}
      & \makecell[br]{Output\\Tokens} \\
    \midrule
    \multirow{3}{*}{\textit{Getting Started}}
      & \megatron   & 40.5 &  5.4 &  88 &  69.5K & 26.9K \\
      & \torchtitan & 11.4 &  5.2 &  54 &  56.8K & 15.8K \\
      & \sys        & \textbf{6.6} & \textbf{3.1} & \textbf{26} & \textbf{37.0K} & \textbf{5.8K} \\
    \midrule
    \multirow{3}{*}{\textit{Train and Evaluate}}
      & \megatron   & 55.5 & 36.0 & 163 & 106.4K & 52.9K \\
      & \torchtitan & 72.5 & 36.3 & 212 & 176.6K & 97.8K \\
      & \sys        & \textbf{38.5} & \textbf{22.7} & \textbf{92} & \textbf{85.5K} & \textbf{34.2K} \\
    \midrule
    \multirow{3}{*}{\textit{Collect Routing Trace}}
      & \megatron   & 33.3 &  5.5 & 112 & 144.3K & 102.1K \\
      & \torchtitan & 32.8 & 10.4 & 103 & 166.0K &  84.7K \\
      & \sys        & \textbf{16.3} & \textbf{2.8} & \textbf{58} & \textbf{118.7K} & \textbf{56.2K} \\
    \midrule
    \multirow{3}{*}{\textit{Report Heavy Kernels}}
      & \megatron   & 22.1 & 12.1 & 60 & 52.6K & 23.9K \\
      & \torchtitan & 15.0 &  6.7 & \textbf{40} & 66.1K & 22.5K \\
      & \sys        & \textbf{11.8} & \textbf{3.6} & 42 & \textbf{49.2K} & \textbf{16.0K} \\
    \bottomrule
  \end{tabular}
\end{table}

\MyPara{New Feature.}
New-feature tasks exercise the test--debug cycle: edit, run
training, read crash, edit again.
{Across new-feature tasks (\autoref{tab:agent-efforts}), \sys's
Active GPU Time is up to 44\% lower than \megatron's and 64\% lower
than \torchtitan's. Debugging \torchtitan's OOM failures on MoBA
contributes to the latter gap. Excluding tasks where \torchtitan runs
out of memory, the maximum reduction is 32\% on Diff.}
Two patterns inflate \megatron's reruns: a hidden argument
registry causes the agent's manually-added CLI flags to collide
with auto-derived ones (implicit indirection), and C++ paths like
TransformerEngine's grouped-GEMM emit opaque segfaults that drive
speculative configuration toggles (not Python-native).
\torchtitan's reruns are dominated by memory-pressure debugging.
On \sys, failures surface inside the file the agent just wrote
with a readable Python traceback, and fixes stay in the same
file. \S\ref{sec:eval:case} provides a detailed case study on
MoBA.

\begin{table}[t]
  \centering
  \small
  \vspace{-5pt}
  \caption{Per-task agent effort across frameworks on new-feature
  tasks (\S\ref{sec:benchmark}). Each metric reports the median of
  three independent attempts. Lower is more efficient. Session Duration
  and Active GPU Time are in minutes. Per-attempt metrics in
  {Appendix~\ref{sec:appendix:claude}}.
  }
  \label{tab:agent-efforts}
  \setlength{\tabcolsep}{3pt}
  \begin{tabular}{@{}lr|rrrrr@{}}
    \toprule
    Task
      & Framework
      & \makecell[br]{Session\\Duration}
      & \makecell[br]{Active\\GPU Time}
      & \makecell[br]{Agent\\Turns}
      & \makecell[br]{Per-Turn\\Context}
      & \makecell[br]{Output\\Tokens} \\
    \midrule
    \multirow{3}{*}{\makecell[l]{\textit{Differential Transformer}\\(Diff)~\cite{ye2024differential}}}
      & \megatron   & 47.1 & 33.7 & 125 & 118.7K & 57.1K \\
      & \torchtitan & 49.6 & 40.3 &  58 & 103.2K & 36.0K \\
      & \sys        & \textbf{38.2} & \textbf{27.6} & \textbf{47} & \textbf{69.5K} & \textbf{25.4K} \\
    \midrule
    \multirow{3}{*}{\makecell[l]{\textit{Dynamic Mixture of Experts}\\(DynMoE)~\cite{guo2024dynamic}}}
      & \megatron   &  83.8 & 49.1 & 199 & 208.0K & 115.2K \\
      & \torchtitan & 140.6 & 94.4 & 197 & 228.8K & 161.3K \\
      & \sys        & \textbf{60.4} & \textbf{41.9} & \textbf{76} & \textbf{146.0K} & \textbf{76.4K} \\
    \midrule
    \multirow{3}{*}{\makecell[l]{\textit{Mixture of Block Attention}\\(MoBA)~\cite{lu2025moba}}}
      & \megatron   &  61.6 & 49.5 & 134 & 120.9K &  53.8K \\
      & \torchtitan & 105.1 & 77.9 &  91 & 166.9K & 111.8K \\
      & \sys        & \textbf{38.7} & \textbf{27.7} & \textbf{57} & \textbf{69.0K} & \textbf{32.4K} \\
    \midrule
    \multirow{3}{*}{\textit{MoE++}~\cite{jin2025moeplusplus}}
      & \megatron   & 88.5 & 58.7 & 145 & 188.7K & 117.0K \\
      & \torchtitan & 71.4 & 51.9 & \textbf{87} & \textbf{164.6K} & \textbf{85.3K} \\
      & \sys        & \textbf{63.0} & \textbf{39.9} &  90 & 176.6K & 107.7K \\
    \bottomrule
  \end{tabular}
\end{table}

\subsection{Ablation Study on Agent Skills}
\label{sec:eval:skills-ablation}

In this section, we isolate the effect of agent skills via
ablation. They are a self-contained set of files shipped in the
repository, so we can toggle them on and off against an otherwise
fixed codebase. We
pick two of
\sys's skills, \texttt{validate-correctness} and
\texttt{capture-nsys-profile}, which mirror the natural follow-up
after a system optimization: validate that training correctness is
preserved, then capture an Nsight Systems profile to examine
whether the optimization is effective.

We run this ablation on the wgrad delay~\cite{qi2024zero} commit in
\sys, repeating each task three times with skills and three times
without. The codebase, agent, and harness are otherwise identical,
and we report the same five effort metrics as in
\S\ref{sec:eval:agent}. When skills are disabled, we strip them
from both the working tree and the git history, so the agent cannot
recover the procedure from either. All twelve runs completed
successfully, reporting the correct verdict or generating a valid
Nsight Systems profile.

\autoref{tab:skills-ablation} reports the results. Active GPU
Time stays near parity across both tasks: each task runs a fixed
set of training runs pinned by the workflow, so the GPU work is
determined by the task rather than the agent. The four agent-side
metrics, which capture the agent's reasoning overhead in setup,
launch, and interpretation, all drop substantially with skills
enabled. Agent turns drop the most (70\% and 52\% respectively),
suggesting that with the procedure encoded in the skill, the agent
acts on a fixed plan rather than iteratively deriving one through
repeated tool calls. These results demonstrate that task-specific
in-repo skills, comprising the markdown playbook and any helper
scripts, substantially reduce agent effort on the recurring
training-system tasks they target. {We also evaluate profiling
skills on \megatron and \torchtitan in
Appendix~\ref{sec:appendix:baseline-skills}.}

\begin{table}[t]
  \centering
  \small
  \caption{Agent effort on \sys with and without the corresponding
  skill. Each metric reports the median of three independent
  attempts. Session Duration and Active GPU Time are in minutes.}
  \label{tab:skills-ablation}
  \setlength{\tabcolsep}{3pt}
  \begin{tabular}{@{}lr|rrrrr@{}}
    \toprule
    Task
      & Skills
      & \makecell[br]{Session\\Duration}
      & \makecell[br]{Active\\GPU Time}
      & \makecell[br]{Agent\\Turns}
      & \makecell[br]{Per-Turn\\Context}
      & \makecell[br]{Output\\Tokens} \\
    \midrule
    \multirow{2}{*}{\texttt{validate-correctness}}
      & off & 26.0 & 20.8 & 114 & 96.3K & 30.2K \\
      & on  & 22.9 & 22.5 &  34 & 43.5K & 11.3K \\
    \midrule
    \multirow{2}{*}{\texttt{capture-nsys-profile}}
      & off &  9.4 &  5.6 &  75 & 62.1K & 25.3K \\
      & on  &  6.6 &  5.5 &  36 & 40.3K & 14.5K \\
    \bottomrule
  \end{tabular}
\end{table}

\subsection{Case Study: Integrating MoBA}
\label{sec:eval:case}

\begin{figure}[t]
  \centering
  \begin{subfigure}[t]{0.45\linewidth}
    \centering
    \includegraphics[width=\linewidth]{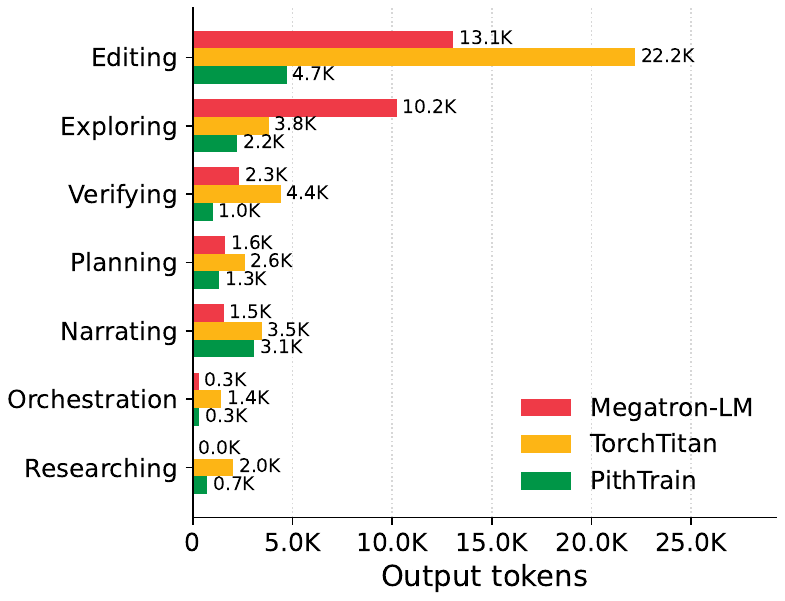}
    \caption{Per-category breakdown of output tokens.}
    \label{fig:moba-token-breakdown}
  \end{subfigure}
  \hfill
  \begin{subfigure}[t]{0.45\linewidth}
    \centering
    \includegraphics[width=\linewidth]{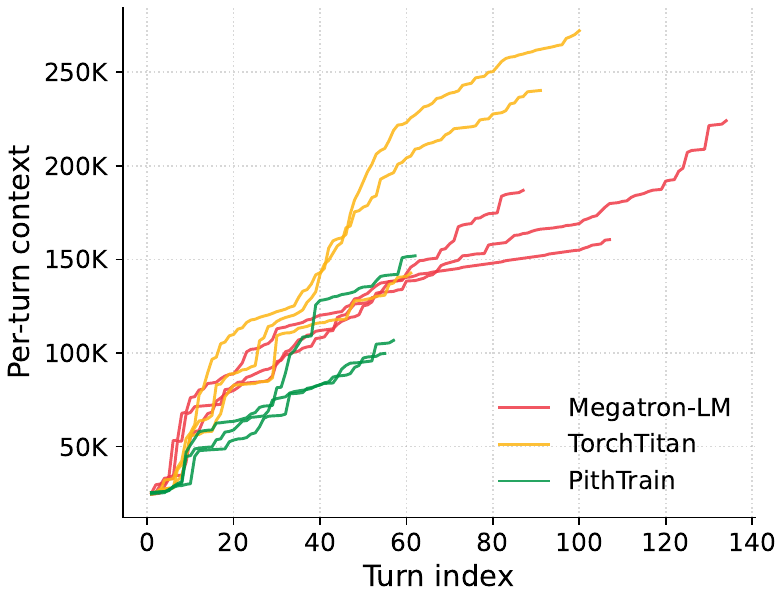}
    \caption{Per-turn context window over the session.}
    \label{fig:moba-context-trace}
  \end{subfigure}
  \caption{Agent behavior on integrating MoBA across frameworks.
  (\subref{fig:moba-token-breakdown}) reports the median output
  tokens per action category of three independent attempts;
  (\subref{fig:moba-context-trace}) shows the per-turn input-side
  context for each of three independent attempts.}
  \label{fig:moba-case-figures}
\end{figure}

To examine where per-framework differences in agent cost come from,
we conduct a case study on integrating MoBA, decomposing the
agent's output tokens by action
category\footnote{Thinking tokens are excluded as they are not
surfaced by the API.}~(\autoref{fig:moba-token-breakdown}) and
tracing its per-turn context
window~(\autoref{fig:moba-context-trace}). Editing dominates
across all three frameworks (\sys 4.7K, \megatron 13.1K,
\torchtitan 22.2K). \megatron also spends substantially more on
Exploring (10.2K vs.\ 3.8K for \torchtitan and 2.2K for \sys),
and its per-turn context sits well above \sys's. The agent reads
the codebase to locate edits and interpret tracebacks, so a
compact codebase with no implicit indirection lowers both
Exploring and per-turn context. \torchtitan's elevated Editing
and \textasciitilde2$\times$ context spike in two of three runs have a different
cause: the agent's initial implementation runs out of memory
(OOM), forcing repeated debug-edit cycles. This points to runtime
properties like memory headroom as a factor independent of
codebase structure.

Beyond \torchtitan's memory failures, we examine the other
failures the agent encountered. Across the three \sys runs, two
complete without any failure; the third hits a tensor-stride
mismatch in the agent's custom attention kernel, fixed in the same
file as the traceback. \megatron's three runs hit two distinct failures: two runs fail
with a duplicate command-line flag registration that conflicts with
one defined in framework code, and one run fails with a BF16
overflow in the agent's code. Each fix on \megatron spans multiple
files. This
contrast reflects \sys's compactness and absence of implicit
indirection, which keep each fix local to the agent's edit.

\par\begingroup
\section{Discussion and Limitations}\label{sec:limitations}
The four principles interact, and our comparisons evaluate their
combined effect. \sysbench does not yet cover propagating shared
changes across models, where implicit indirection may lower agent effort.
Maintaining ATE as a framework grows is another open direction. For
example, we use agent reviewers to flag compactness concerns and encourage
agents to refine skills after use. Periodic \sysbench evaluation and
focused model coverage could further help manage growth. We hope \sys
and \sysbench help the community explore how to preserve agent-task
efficiency as training frameworks evolve.
\par\endgroup

\section{Conclusion}
\label{sec:conclusion}

We presented \sys, a compact, agent-native MoE training framework
built on four design principles.
\sys{} {offers competitive throughput} across a
range of models, and on \sysbench, a coding agent achieves higher
agent-task efficiency on \sys than on production frameworks.
We hope \sys serves as a starting point for future agent-native
training {frameworks}.

\begin{ack}
{This work is supported in part by the National Science Foundation under
grant number IIS2543593, funding from Bosch, Amazon, and gifts from NVIDIA and Google.}
{We thank the CMU Foundation and Language Model (FLAME) Center for
providing compute resources to develop \sys, and NVIDIA for DGX B200 support.}
{We also thank Zichu Wu, Gaokai Zhang, Rishith Prathi, and Yagneek Patlola for their
contributions to the ongoing development of \sys.}
\end{ack}

\bibliography{reference}

@misc{deepseekai2025deepseekv3technicalreport,
      title={DeepSeek-V3 Technical Report}, 
      author={DeepSeek-AI and Aixin Liu and Bei Feng and Bing Xue and Bingxuan Wang and Bochao Wu and Chengda Lu and Chenggang Zhao and Chengqi Deng and Chenyu Zhang and Chong Ruan and Damai Dai and Daya Guo and Dejian Yang and Deli Chen and Dongjie Ji and Erhang Li and Fangyun Lin and Fucong Dai and Fuli Luo and Guangbo Hao and Guanting Chen and Guowei Li and H. Zhang and Han Bao and Hanwei Xu and Haocheng Wang and Haowei Zhang and Honghui Ding and Huajian Xin and Huazuo Gao and Hui Li and Hui Qu and J. L. Cai and Jian Liang and Jianzhong Guo and Jiaqi Ni and Jiashi Li and Jiawei Wang and Jin Chen and Jingchang Chen and Jingyang Yuan and Junjie Qiu and Junlong Li and Junxiao Song and Kai Dong and Kai Hu and Kaige Gao and Kang Guan and Kexin Huang and Kuai Yu and Lean Wang and Lecong Zhang and Lei Xu and Leyi Xia and Liang Zhao and Litong Wang and Liyue Zhang and Meng Li and Miaojun Wang and Mingchuan Zhang and Minghua Zhang and Minghui Tang and Mingming Li and Ning Tian and Panpan Huang and Peiyi Wang and Peng Zhang and Qiancheng Wang and Qihao Zhu and Qinyu Chen and Qiushi Du and R. J. Chen and R. L. Jin and Ruiqi Ge and Ruisong Zhang and Ruizhe Pan and Runji Wang and Runxin Xu and Ruoyu Zhang and Ruyi Chen and S. S. Li and Shanghao Lu and Shangyan Zhou and Shanhuang Chen and Shaoqing Wu and Shengfeng Ye and Shengfeng Ye and Shirong Ma and Shiyu Wang and Shuang Zhou and Shuiping Yu and Shunfeng Zhou and Shuting Pan and T. Wang and Tao Yun and Tian Pei and Tianyu Sun and W. L. Xiao and Wangding Zeng and Wanjia Zhao and Wei An and Wen Liu and Wenfeng Liang and Wenjun Gao and Wenqin Yu and Wentao Zhang and X. Q. Li and Xiangyue Jin and Xianzu Wang and Xiao Bi and Xiaodong Liu and Xiaohan Wang and Xiaojin Shen and Xiaokang Chen and Xiaokang Zhang and Xiaosha Chen and Xiaotao Nie and Xiaowen Sun and Xiaoxiang Wang and Xin Cheng and Xin Liu and Xin Xie and Xingchao Liu and Xingkai Yu and Xinnan Song and Xinxia Shan and Xinyi Zhou and Xinyu Yang and Xinyuan Li and Xuecheng Su and Xuheng Lin and Y. K. Li and Y. Q. Wang and Y. X. Wei and Y. X. Zhu and Yang Zhang and Yanhong Xu and Yanhong Xu and Yanping Huang and Yao Li and Yao Zhao and Yaofeng Sun and Yaohui Li and Yaohui Wang and Yi Yu and Yi Zheng and Yichao Zhang and Yifan Shi and Yiliang Xiong and Ying He and Ying Tang and Yishi Piao and Yisong Wang and Yixuan Tan and Yiyang Ma and Yiyuan Liu and Yongqiang Guo and Yu Wu and Yuan Ou and Yuchen Zhu and Yuduan Wang and Yue Gong and Yuheng Zou and Yujia He and Yukun Zha and Yunfan Xiong and Yunxian Ma and Yuting Yan and Yuxiang Luo and Yuxiang You and Yuxuan Liu and Yuyang Zhou and Z. F. Wu and Z. Z. Ren and Zehui Ren and Zhangli Sha and Zhe Fu and Zhean Xu and Zhen Huang and Zhen Zhang and Zhenda Xie and Zhengyan Zhang and Zhewen Hao and Zhibin Gou and Zhicheng Ma and Zhigang Yan and Zhihong Shao and Zhipeng Xu and Zhiyu Wu and Zhongyu Zhang and Zhuoshu Li and Zihui Gu and Zijia Zhu and Zijun Liu and Zilin Li and Ziwei Xie and Ziyang Song and Ziyi Gao and Zizheng Pan},
      year={2025},
      eprint={2412.19437},
      archivePrefix={arXiv},
      primaryClass={cs.CL},
      url={https://arxiv.org/abs/2412.19437}, 
}

@misc{yang2025qwen3technicalreport,
      title={Qwen3 Technical Report}, 
      author={An Yang and Anfeng Li and Baosong Yang and Beichen Zhang and Binyuan Hui and Bo Zheng and Bowen Yu and Chang Gao and Chengen Huang and Chenxu Lv and Chujie Zheng and Dayiheng Liu and Fan Zhou and Fei Huang and Feng Hu and Hao Ge and Haoran Wei and Huan Lin and Jialong Tang and Jian Yang and Jianhong Tu and Jianwei Zhang and Jianxin Yang and Jiaxi Yang and Jing Zhou and Jingren Zhou and Junyang Lin and Kai Dang and Keqin Bao and Kexin Yang and Le Yu and Lianghao Deng and Mei Li and Mingfeng Xue and Mingze Li and Pei Zhang and Peng Wang and Qin Zhu and Rui Men and Ruize Gao and Shixuan Liu and Shuang Luo and Tianhao Li and Tianyi Tang and Wenbiao Yin and Xingzhang Ren and Xinyu Wang and Xinyu Zhang and Xuancheng Ren and Yang Fan and Yang Su and Yichang Zhang and Yinger Zhang and Yu Wan and Yuqiong Liu and Zekun Wang and Zeyu Cui and Zhenru Zhang and Zhipeng Zhou and Zihan Qiu},
      year={2025},
      eprint={2505.09388},
      archivePrefix={arXiv},
      primaryClass={cs.CL},
      url={https://arxiv.org/abs/2505.09388}, 
}

@misc{openai2025gptoss120bgptoss20bmodel,
      title={{gpt-oss-120b} \& {gpt-oss-20b} {Model Card}},
      author={OpenAI and : and Sandhini Agarwal and Lama Ahmad and Jason Ai and Sam Altman and Andy Applebaum and Edwin Arbus and Rahul K. Arora and Yu Bai and Bowen Baker and Haiming Bao and Boaz Barak and Ally Bennett and Tyler Bertao and Nivedita Brett and Eugene Brevdo and Greg Brockman and Sebastien Bubeck and Che Chang and Kai Chen and Mark Chen and Enoch Cheung and Aidan Clark and Dan Cook and Marat Dukhan and Casey Dvorak and Kevin Fives and Vlad Fomenko and Timur Garipov and Kristian Georgiev and Mia Glaese and Tarun Gogineni and Adam Goucher and Lukas Gross and Katia Gil Guzman and John Hallman and Jackie Hehir and Johannes Heidecke and Alec Helyar and Haitang Hu and Romain Huet and Jacob Huh and Saachi Jain and Zach Johnson and Chris Koch and Irina Kofman and Dominik Kundel and Jason Kwon and Volodymyr Kyrylov and Elaine Ya Le and Guillaume Leclerc and James Park Lennon and Scott Lessans and Mario Lezcano-Casado and Yuanzhi Li and Zhuohan Li and Ji Lin and Jordan Liss and Lily and Liu and Jiancheng Liu and Kevin Lu and Chris Lu and Zoran Martinovic and Lindsay McCallum and Josh McGrath and Scott McKinney and Aidan McLaughlin and Song Mei and Steve Mostovoy and Tong Mu and Gideon Myles and Alexander Neitz and Alex Nichol and Jakub Pachocki and Alex Paino and Dana Palmie and Ashley Pantuliano and Giambattista Parascandolo and Jongsoo Park and Leher Pathak and Carolina Paz and Ludovic Peran and Dmitry Pimenov and Michelle Pokrass and Elizabeth Proehl and Huida Qiu and Gaby Raila and Filippo Raso and Hongyu Ren and Kimmy Richardson and David Robinson and Bob Rotsted and Hadi Salman and Suvansh Sanjeev and Max Schwarzer and D. Sculley and Harshit Sikchi and Kendal Simon and Karan Singhal and Yang Song and Dane Stuckey and Zhiqing Sun and Philippe Tillet and Sam Toizer and Foivos Tsimpourlas and Nikhil Vyas and Eric Wallace and Xin Wang and Miles Wang and Olivia Watkins and Kevin Weil and Amy Wendling and Kevin Whinnery and Cedric Whitney and Hannah Wong and Lin Yang and Yu Yang and Michihiro Yasunaga and Kristen Ying and Wojciech Zaremba and Wenting Zhan and Cyril Zhang and Brian Zhang and Eddie Zhang and Shengjia Zhao},
      year={2025},
      eprint={2508.10925},
      archivePrefix={arXiv},
      primaryClass={cs.CL},
      url={https://arxiv.org/abs/2508.10925}, 
}

@misc{deepseekai2024deepseekv2strongeconomicalefficient,
      title={DeepSeek-V2: A Strong, Economical, and Efficient Mixture-of-Experts Language Model}, 
      author={DeepSeek-AI and Aixin Liu and Bei Feng and Bin Wang and Bingxuan Wang and Bo Liu and Chenggang Zhao and Chengqi Dengr and Chong Ruan and Damai Dai and Daya Guo and Dejian Yang and Deli Chen and Dongjie Ji and Erhang Li and Fangyun Lin and Fuli Luo and Guangbo Hao and Guanting Chen and Guowei Li and H. Zhang and Hanwei Xu and Hao Yang and Haowei Zhang and Honghui Ding and Huajian Xin and Huazuo Gao and Hui Li and Hui Qu and J. L. Cai and Jian Liang and Jianzhong Guo and Jiaqi Ni and Jiashi Li and Jin Chen and Jingyang Yuan and Junjie Qiu and Junxiao Song and Kai Dong and Kaige Gao and Kang Guan and Lean Wang and Lecong Zhang and Lei Xu and Leyi Xia and Liang Zhao and Liyue Zhang and Meng Li and Miaojun Wang and Mingchuan Zhang and Minghua Zhang and Minghui Tang and Mingming Li and Ning Tian and Panpan Huang and Peiyi Wang and Peng Zhang and Qihao Zhu and Qinyu Chen and Qiushi Du and R. J. Chen and R. L. Jin and Ruiqi Ge and Ruizhe Pan and Runxin Xu and Ruyi Chen and S. S. Li and Shanghao Lu and Shangyan Zhou and Shanhuang Chen and Shaoqing Wu and Shengfeng Ye and Shirong Ma and Shiyu Wang and Shuang Zhou and Shuiping Yu and Shunfeng Zhou and Size Zheng and T. Wang and Tian Pei and Tian Yuan and Tianyu Sun and W. L. Xiao and Wangding Zeng and Wei An and Wen Liu and Wenfeng Liang and Wenjun Gao and Wentao Zhang and X. Q. Li and Xiangyue Jin and Xianzu Wang and Xiao Bi and Xiaodong Liu and Xiaohan Wang and Xiaojin Shen and Xiaokang Chen and Xiaosha Chen and Xiaotao Nie and Xiaowen Sun and Xiaoxiang Wang and Xin Liu and Xin Xie and Xingkai Yu and Xinnan Song and Xinyi Zhou and Xinyu Yang and Xuan Lu and Xuecheng Su and Y. Wu and Y. K. Li and Y. X. Wei and Y. X. Zhu and Yanhong Xu and Yanping Huang and Yao Li and Yao Zhao and Yaofeng Sun and Yaohui Li and Yaohui Wang and Yi Zheng and Yichao Zhang and Yiliang Xiong and Yilong Zhao and Ying He and Ying Tang and Yishi Piao and Yixin Dong and Yixuan Tan and Yiyuan Liu and Yongji Wang and Yongqiang Guo and Yuchen Zhu and Yuduan Wang and Yuheng Zou and Yukun Zha and Yunxian Ma and Yuting Yan and Yuxiang You and Yuxuan Liu and Z. Z. Ren and Zehui Ren and Zhangli Sha and Zhe Fu and Zhen Huang and Zhen Zhang and Zhenda Xie and Zhewen Hao and Zhihong Shao and Zhiniu Wen and Zhipeng Xu and Zhongyu Zhang and Zhuoshu Li and Zihan Wang and Zihui Gu and Zilin Li and Ziwei Xie},
      year={2024},
      eprint={2405.04434},
      archivePrefix={arXiv},
      primaryClass={cs.CL},
      url={https://arxiv.org/abs/2405.04434}, 
}

@inproceedings{
lepikhin2020gshardscalinggiantmodels,
title={{\{}GS{\}}hard: Scaling Giant Models with Conditional Computation and Automatic Sharding},
author={Dmitry Lepikhin and HyoukJoong Lee and Yuanzhong Xu and Dehao Chen and Orhan Firat and Yanping Huang and Maxim Krikun and Noam Shazeer and Zhifeng Chen},
booktitle={International Conference on Learning Representations},
year={2021},
url={https://openreview.net/forum?id=qrwe7XHTmYb}
}

@misc{shoeybi2020megatronlmtrainingmultibillionparameter,
      title={Megatron-LM: Training Multi-Billion Parameter Language Models Using Model Parallelism}, 
      author={Mohammad Shoeybi and Mostofa Patwary and Raul Puri and Patrick LeGresley and Jared Casper and Bryan Catanzaro},
      year={2020},
      eprint={1909.08053},
      archivePrefix={arXiv},
      primaryClass={cs.CL},
      url={https://arxiv.org/abs/1909.08053}, 
}

@inproceedings{narayanan2021efficientlargescalelanguagemodel,
author = {Narayanan, Deepak and Shoeybi, Mohammad and Casper, Jared and LeGresley, Patrick and Patwary, Mostofa and Korthikanti, Vijay and Vainbrand, Dmitri and Kashinkunti, Prethvi and Bernauer, Julie and Catanzaro, Bryan and Phanishayee, Amar and Zaharia, Matei},
title = {Efficient large-scale language model training on GPU clusters using megatron-LM},
year = {2021},
isbn = {9781450384421},
publisher = {Association for Computing Machinery},
address = {New York, NY, USA},
url = {https://doi.org/10.1145/3458817.3476209},
doi = {10.1145/3458817.3476209},
booktitle = {Proceedings of the International Conference for High Performance Computing, Networking, Storage and Analysis},
articleno = {58},
numpages = {15},
location = {St. Louis, Missouri},
series = {SC '21}
}

@inproceedings{deepspeed,
author = {Rasley, Jeff and Rajbhandari, Samyam and Ruwase, Olatunji and He, Yuxiong},
title = {DeepSpeed: System Optimizations Enable Training Deep Learning Models with Over 100 Billion Parameters},
year = {2020},
isbn = {9781450379984},
publisher = {Association for Computing Machinery},
address = {New York, NY, USA},
url = {https://doi.org/10.1145/3394486.3406703},
doi = {10.1145/3394486.3406703},
booktitle = {Proceedings of the 26th ACM SIGKDD International Conference on Knowledge Discovery \& Data Mining},
pages = {3505–3506},
numpages = {2},
location = {Virtual Event, CA, USA},
series = {KDD '20}
}

@inproceedings{rajbhandari2020zeromemoryoptimizationstraining,
author = {Rajbhandari, Samyam and Rasley, Jeff and Ruwase, Olatunji and He, Yuxiong},
title = {ZeRO: memory optimizations toward training trillion parameter models},
year = {2020},
isbn = {9781728199986},
publisher = {IEEE Press},
booktitle = {Proceedings of the International Conference for High Performance Computing, Networking, Storage and Analysis},
articleno = {20},
numpages = {16},
location = {Atlanta, Georgia},
series = {SC '20}
}

@inproceedings{
liang2025torchtitanonestoppytorchnative,
title={TorchTitan: One-stop PyTorch native solution for production ready {LLM} pretraining},
author={Wanchao Liang and Tianyu Liu and Less Wright and Will Constable and Andrew Gu and Chien-Chin Huang and Iris Zhang and Wei Feng and Howard Huang and Junjie Wang and Sanket Purandare and Gokul Nadathur and Stratos Idreos},
booktitle={The Thirteenth International Conference on Learning Representations},
year={2025},
url={https://openreview.net/forum?id=SFN6Wm7YBI}
}

@misc{claude-code,                                              
  author       = {{Anthropic}},
  title        = {{Claude Code}},                           
  year         = {2025},       
  howpublished = {\url{https://www.anthropic.com/claude-code}}
}

@misc{codex-cli,                                                                                                                   
  author       = {{OpenAI}},
  title        = {{Codex CLI}},                                                                                                    
  year         = {2025},       
  howpublished = {\url{https://github.com/openai/codex}}        
}

@misc
{
    cursor,
    author = {{Anysphere}},
    title = {{Cursor: The AI code editor}},
    year = {2023},
    howpublished = {\url{https://www.cursor.com}}
}

@misc{github-copilot,                                                                                                              
  author       = {{GitHub}},
  title        = {{GitHub Copilot}},
  year         = {2021},       
  howpublished = {\url{https://github.com/features/copilot}}                                                                       
}

@inproceedings{
yang2024sweagentagentcomputerinterfacesenable,
title={{SWE}-agent: Agent-Computer Interfaces Enable Automated Software Engineering},
author={John Yang and Carlos E Jimenez and Alexander Wettig and Kilian Lieret and Shunyu Yao and Karthik R Narasimhan and Ofir Press},
booktitle={The Thirty-eighth Annual Conference on Neural Information Processing Systems},
year={2024},
url={https://openreview.net/forum?id=mXpq6ut8J3}
}

@inproceedings{zhang2024autocoderoverautonomousprogramimprovement,
author = {Zhang, Yuntong and Ruan, Haifeng and Fan, Zhiyu and Roychoudhury, Abhik},
title = {AutoCodeRover: Autonomous Program Improvement},
year = {2024},
isbn = {9798400706127},
publisher = {Association for Computing Machinery},
address = {New York, NY, USA},
url = {https://doi.org/10.1145/3650212.3680384},
doi = {10.1145/3650212.3680384},
booktitle = {Proceedings of the 33rd ACM SIGSOFT International Symposium on Software Testing and Analysis},
pages = {1592–1604},
numpages = {13},
location = {Vienna, Austria},
series = {ISSTA 2024}
}

@inproceedings{
jimenez2024swebench,
title={{SWE}-bench: Can Language Models Resolve Real-world Github Issues?},
author={Carlos E Jimenez and John Yang and Alexander Wettig and Shunyu Yao and Kexin Pei and Ofir Press and Karthik R Narasimhan},
booktitle={The Twelfth International Conference on Learning Representations},
year={2024},
url={https://openreview.net/forum?id=VTF8yNQM66}
}

@inproceedings{
chan2025mlebench,
title={{MLE}-bench: Evaluating Machine Learning Agents on Machine Learning Engineering},
author={Jun Shern Chan and Neil Chowdhury and Oliver Jaffe and James Aung and Dane Sherburn and Evan Mays and Giulio Starace and Kevin Liu and Leon Maksin and Tejal Patwardhan and Aleksander Madry and Lilian Weng},
booktitle={The Thirteenth International Conference on Learning Representations},
year={2025},
url={https://openreview.net/forum?id=6s5uXNWGIh}
}

@misc{chen2021evaluatinglargelanguagemodels,
      title={Evaluating Large Language Models Trained on Code}, 
      author={Mark Chen and Jerry Tworek and Heewoo Jun and Qiming Yuan and Henrique Ponde de Oliveira Pinto and Jared Kaplan and Harri Edwards and Yuri Burda and Nicholas Joseph and Greg Brockman and Alex Ray and Raul Puri and Gretchen Krueger and Michael Petrov and Heidy Khlaaf and Girish Sastry and Pamela Mishkin and Brooke Chan and Scott Gray and Nick Ryder and Mikhail Pavlov and Alethea Power and Lukasz Kaiser and Mohammad Bavarian and Clemens Winter and Philippe Tillet and Felipe Petroski Such and Dave Cummings and Matthias Plappert and Fotios Chantzis and Elizabeth Barnes and Ariel Herbert-Voss and William Hebgen Guss and Alex Nichol and Alex Paino and Nikolas Tezak and Jie Tang and Igor Babuschkin and Suchir Balaji and Shantanu Jain and William Saunders and Christopher Hesse and Andrew N. Carr and Jan Leike and Josh Achiam and Vedant Misra and Evan Morikawa and Alec Radford and Matthew Knight and Miles Brundage and Mira Murati and Katie Mayer and Peter Welinder and Bob McGrew and Dario Amodei and Sam McCandlish and Ilya Sutskever and Wojciech Zaremba},
      year={2021},
      eprint={2107.03374},
      archivePrefix={arXiv},
      primaryClass={cs.LG},
      url={https://arxiv.org/abs/2107.03374}, 
}

@misc{austin2021programsynthesislargelanguage,
      title={Program Synthesis with Large Language Models}, 
      author={Jacob Austin and Augustus Odena and Maxwell Nye and Maarten Bosma and Henryk Michalewski and David Dohan and Ellen Jiang and Carrie Cai and Michael Terry and Quoc Le and Charles Sutton},
      year={2021},
      eprint={2108.07732},
      archivePrefix={arXiv},
      primaryClass={cs.PL},
      url={https://arxiv.org/abs/2108.07732}, 
}

@article{pytorchfsdp,
author = {Zhao, Yanli and Gu, Andrew and Varma, Rohan and Luo, Liang and Huang, Chien-Chin and Xu, Min and Wright, Less and Shojanazeri, Hamid and Ott, Myle and Shleifer, Sam and Desmaison, Alban and Balioglu, Can and Damania, Pritam and Nguyen, Bernard and Chauhan, Geeta and Hao, Yuchen and Mathews, Ajit and Li, Shen},
title = {PyTorch FSDP: Experiences on Scaling Fully Sharded Data Parallel},
year = {2023},
issue_date = {August 2023},
publisher = {VLDB Endowment},
volume = {16},
number = {12},
issn = {2150-8097},
url = {https://doi.org/10.14778/3611540.3611569},
doi = {10.14778/3611540.3611569},
journal = {Proc. VLDB Endow.},
month = aug,
pages = {3848–3860},
numpages = {13}
}

@inproceedings{
liu2024ringattention,
title={RingAttention with Blockwise Transformers for Near-Infinite Context},
author={Hao Liu and Matei Zaharia and Pieter Abbeel},
booktitle={The Twelfth International Conference on Learning Representations},
year={2024},
url={https://openreview.net/forum?id=WsRHpHH4s0}
}

@misc{micikevicius2022fp8formatsdeeplearning,
      title={FP8 Formats for Deep Learning}, 
      author={Paulius Micikevicius and Dusan Stosic and Neil Burgess and Marius Cornea and Pradeep Dubey and Richard Grisenthwaite and Sangwon Ha and Alexander Heinecke and Patrick Judd and John Kamalu and Naveen Mellempudi and Stuart Oberman and Mohammad Shoeybi and Michael Siu and Hao Wu},
      year={2022},
      eprint={2209.05433},
      archivePrefix={arXiv},
      primaryClass={cs.LG},
      url={https://arxiv.org/abs/2209.05433}, 
}

@inproceedings{
qi2024zero,
title={Zero Bubble (Almost) Pipeline Parallelism},
author={Penghui Qi and Xinyi Wan and Guangxing Huang and Min Lin},
booktitle={The Twelfth International Conference on Learning Representations},
year={2024},
url={https://openreview.net/forum?id=tuzTN0eIO5}
}

@misc{torch-dcp,                       
author       = {{The PyTorch Team}},                                                                                             
title        = {{torch.distributed.checkpoint}},
year         = {2023},                                                                                                           
howpublished = {\url{https://docs.pytorch.org/docs/stable/distributed.checkpoint.html}}
}

@misc{transformer-engine,                                                                                                          
author       = {{NVIDIA}},                                                                                                       
title        = {{TransformerEngine}},
year         = {2022},                                                                                                           
howpublished = {\url{https://github.com/NVIDIA/TransformerEngine}}                                                               
}

@misc
{
    anthropic-skills,
    author = {{Anthropic}},
    title  = {{Equipping agents for the real world with Agent Skills}},
    year   = {2025},
    howpublished = {\url{https://www.anthropic.com/engineering/equipping-agents-for-the-real-world-with-agent-skills}}
}

@inproceedings{
jin2025moeplusplus,
title={MoE++: Accelerating Mixture-of-Experts Methods with Zero-Computation Experts},
author={Peng Jin and Bo Zhu and Li Yuan and Shuicheng YAN},
booktitle={The Thirteenth International Conference on Learning Representations},
year={2025},
url={https://openreview.net/forum?id=t7P5BUKcYv}
}

@misc{kimiteam2026kimik2openagentic,
      title={Kimi K2: Open Agentic Intelligence}, 
      author={Kimi Team and Yifan Bai and Yiping Bao and Y. Charles and Cheng Chen and Guanduo Chen and Haiting Chen and Huarong Chen and Jiahao Chen and Ningxin Chen and Ruijue Chen and Yanru Chen and Yuankun Chen and Yutian Chen and Zhuofu Chen and Jialei Cui and Hao Ding and Mengnan Dong and Angang Du and Chenzhuang Du and Dikang Du and Yulun Du and Yu Fan and Yichen Feng and Kelin Fu and Bofei Gao and Chenxiao Gao and Hongcheng Gao and Peizhong Gao and Tong Gao and Yuyao Ge and Shangyi Geng and Qizheng Gu and Xinran Gu and Longyu Guan and Haiqing Guo and Jianhang Guo and Xiaoru Hao and Tianhong He and Weiran He and Wenyang He and Yunjia He and Chao Hong and Hao Hu and Yangyang Hu and Zhenxing Hu and Weixiao Huang and Zhiqi Huang and Zihao Huang and Tao Jiang and Zhejun Jiang and Xinyi Jin and Yongsheng Kang and Guokun Lai and Cheng Li and Fang Li and Haoyang Li and Ming Li and Wentao Li and Yang Li and Yanhao Li and Yiwei Li and Zhaowei Li and Zheming Li and Hongzhan Lin and Xiaohan Lin and Zongyu Lin and Chengyin Liu and Chenyu Liu and Hongzhang Liu and Jingyuan Liu and Junqi Liu and Liang Liu and Shaowei Liu and T. Y. Liu and Tianwei Liu and Weizhou Liu and Yangyang Liu and Yibo Liu and Yiping Liu and Yue Liu and Zhengying Liu and Enzhe Lu and Haoyu Lu and Lijun Lu and Yashuo Luo and Shengling Ma and Xinyu Ma and Yingwei Ma and Shaoguang Mao and Jie Mei and Xin Men and Yibo Miao and Siyuan Pan and Yebo Peng and Ruoyu Qin and Zeyu Qin and Bowen Qu and Zeyu Shang and Lidong Shi and Shengyuan Shi and Feifan Song and Jianlin Su and Zhengyuan Su and Lin Sui and Xinjie Sun and Flood Sung and Yunpeng Tai and Heyi Tang and Jiawen Tao and Qifeng Teng and Chaoran Tian and Chensi Wang and Dinglu Wang and Feng Wang and Hailong Wang and Haiming Wang and Jianzhou Wang and Jiaxing Wang and Jinhong Wang and Shengjie Wang and Shuyi Wang and Si Wang and Xinyuan Wang and Yao Wang and Yejie Wang and Yiqin Wang and Yuxin Wang and Yuzhi Wang and Zhaoji Wang and Zhengtao Wang and Zhengtao Wang and Zhexu Wang and Chu Wei and Qianqian Wei and Haoning Wu and Wenhao Wu and Xingzhe Wu and Yuxin Wu and Chenjun Xiao and Jin Xie and Xiaotong Xie and Weimin Xiong and Boyu Xu and Jinjing Xu and L. H. Xu and Lin Xu and Suting Xu and Weixin Xu and Xinran Xu and Yangchuan Xu and Ziyao Xu and Jing Xu and Jing Xu and Junjie Yan and Yuzi Yan and Hao Yang and Xiaofei Yang and Yi Yang and Ying Yang and Zhen Yang and Zhilin Yang and Zonghan Yang and Haotian Yao and Xingcheng Yao and Wenjie Ye and Zhuorui Ye and Bohong Yin and Longhui Yu and Enming Yuan and Hongbang Yuan and Mengjie Yuan and Siyu Yuan and Haobing Zhan and Dehao Zhang and Hao Zhang and Wanlu Zhang and Xiaobin Zhang and Yadong Zhang and Yangkun Zhang and Yichi Zhang and Yizhi Zhang and Yongting Zhang and Yu Zhang and Yutao Zhang and Yutong Zhang and Zheng Zhang and Haotian Zhao and Yikai Zhao and Zijia Zhao and Huabin Zheng and Shaojie Zheng and Longguang Zhong and Jianren Zhou and Xinyu Zhou and Zaida Zhou and Jinguo Zhu and Zhen Zhu and Weiyu Zhuang and Xinxing Zu},
      year={2026},
      eprint={2507.20534},
      archivePrefix={arXiv},
      primaryClass={cs.LG},
      url={https://arxiv.org/abs/2507.20534},
}

@article{chowdhery2023palm,
  title={Palm: Scaling language modeling with pathways},
  author={Chowdhery, Aakanksha and Narang, Sharan and Devlin, Jacob and Bosma, Maarten and Mishra, Gaurav and Roberts, Adam and Barham, Paul and Chung, Hyung Won and Sutton, Charles and Gehrmann, Sebastian and others},
  journal={Journal of machine learning research},
  volume={24},
  number={240},
  pages={1--113},
  year={2023}
}

@article{ye2024differential,
  title={Differential transformer},
  author={Ye, Tianzhu and Dong, Li and Xia, Yuqing and Sun, Yutao and Zhu, Yi and Huang, Gao and Wei, Furu},
  journal={arXiv preprint arXiv:2410.05258},
  year={2024}
}

@article{guo2024dynamic,
  title={Dynamic mixture of experts: An auto-tuning approach for efficient transformer models},
  author={Guo, Yongxin and Cheng, Zhenglin and Tang, Xiaoying and Tu, Zhaopeng and Lin, Tao},
  journal={arXiv preprint arXiv:2405.14297},
  year={2024}
}

@article{lu2025moba,
  title={Moba: Mixture of block attention for long-context llms},
  author={Lu, Enzhe and Jiang, Zhejun and Liu, Jingyuan and Du, Yulun and Jiang, Tao and Hong, Chao and Liu, Shaowei and He, Weiran and Yuan, Enming and Wang, Yuzhi and others},
  journal={arXiv preprint arXiv:2502.13189},
  year={2025}
}

@inproceedings{mihaylov2018openbookqa,
  title={Can a Suit of Armor Conduct Electricity? A New Dataset for Open Book Question Answering},
  author={Mihaylov, Todor and Clark, Peter and Khot, Tushar and Sabharwal, Ashish},
  booktitle={Proceedings of the 2018 Conference on Empirical Methods in Natural Language Processing (EMNLP)},
  year={2018}
}

@article{sakaguchi2020winogrande,
  title={WinoGrande: An Adversarial Winograd Schema Challenge at Scale},
  author={Sakaguchi, Keisuke and Bras, Ronan Le and Bhagavatula, Chandra and Choi, Yejin},
  journal={Communications of the ACM},
  volume={64},
  number={9},
  pages={99--106},
  year={2020}
}

@article{clark2018arc,
  title={Think you have Solved Question Answering? Try ARC, the AI2 Reasoning Challenge},
  author={Clark, Peter and Cowhey, Isaac and Etzioni, Oren and Khot, Tushar and Sabharwal, Ashish and Schoenick, Carissa and Tafjord, Oyvind},
  journal={arXiv preprint arXiv:1803.05457},
  year={2018}
}

@inproceedings{zellers2019hellaswag,
  title={HellaSwag: Can a Machine Really Finish Your Sentence?},
  author={Zellers, Rowan and Holtzman, Ari and Bisk, Yonatan and Farhadi, Ali and Choi, Yejin},
  booktitle={Proceedings of the 57th Annual Meeting of the Association for Computational Linguistics (ACL)},
  year={2019}
}

@inproceedings{bisk2020piqa,
  title={{PIQA}: Reasoning about Physical Commonsense in Natural Language},
  author={Bisk, Yonatan and Zellers, Rowan and Le bras, Ronan and Gao, Jianfeng and Choi, Yejin},
  booktitle={Proceedings of the AAAI Conference on Artificial Intelligence (AAAI)},
  year={2020}
}

@article{qiu2025demonsdetailimplementingload,
  title={Demons in the Detail: On Implementing Load Balancing Loss for Training Specialized Mixture-of-Expert Models},
  author={Qiu, Zihan and Huang, Zeyu and Zheng, Bo and Wen, Kaiyue and Wang, Zekun and Men, Rui and Titov, Ivan and Liu, Dayiheng and Zhou, Jingren and Lin, Junyang},
  journal={arXiv preprint arXiv:2501.11873},
  year={2025}
}

@article{kingma2015adam,
  title={{Adam}: A Method for Stochastic Optimization},
  author={Kingma, Diederik P. and Ba, Jimmy},
  journal={Proceedings of the International Conference on Learning Representations (ICLR)},
  year={2015}
}

@article{li2024dclm,
  title={{DataComp-LM}: In Search of the Next Generation of Training Sets for Language Models},
  author={Li, Jeffrey and Fang, Alex and Smyrnis, Georgios and Ivgi, Maor and Jordan, Matt and Gadre, Samir and Bansal, Hritik and others},
  journal={arXiv preprint arXiv:2406.11794},
  year={2024}
}

@inproceedings{
yao2023react,
title={ReAct: Synergizing Reasoning and Acting in Language Models},
author={Shunyu Yao and Jeffrey Zhao and Dian Yu and Nan Du and Izhak Shafran and Karthik R Narasimhan and Yuan Cao},
booktitle={The Eleventh International Conference on Learning Representations },
year={2023},
url={https://openreview.net/forum?id=WE_vluYUL-X}
}

@misc{xing2026flashinferbenchbuildingvirtuouscycle,
      title={FlashInfer-Bench: Building the Virtuous Cycle for AI-driven LLM Systems}, 
      author={Shanli Xing and Yiyan Zhai and Alexander Jiang and Yixin Dong and Yong Wu and Zihao Ye and Charlie Ruan and Yingyi Huang and Yineng Zhang and Liangsheng Yin and Aksara Bayyapu and Luis Ceze and Tianqi Chen},
      year={2026},
      eprint={2601.00227},
      archivePrefix={arXiv},
      primaryClass={cs.AI},
      url={https://arxiv.org/abs/2601.00227}, 
}

@inproceedings{
ouyang2025kernelbench,
title={KernelBench: Can {LLM}s Write Efficient {GPU} Kernels?},
author={Anne Ouyang and Simon Guo and Simran Arora and Alex L Zhang and William Hu and Christopher Re and Azalia Mirhoseini},
booktitle={Forty-second International Conference on Machine Learning},
year={2025},
url={https://openreview.net/forum?id=yeoN1iQT1x}
}

@misc{ni2025gittaskbench,
      title={{GitTaskBench}: A Benchmark for Code Agents Solving Real-World Tasks Through Code Repository Leveraging}, 
      author={Ziyi Ni and Huacan Wang and Shuo Zhang and Shuo Lu and Ziyang He and Wang You and Zhenheng Tang and Yuntao Du and Bill Sun and Hongzhang Liu and Sen Hu and Ronghao Chen and Bo Li and Xin Li and Chen Hu and Binxing Jiao and Daxin Jiang and Pin Lyu},
      year={2025},
      eprint={2508.18993},
      archivePrefix={arXiv},
      primaryClass={cs.SE},
      url={https://arxiv.org/abs/2508.18993}, 
}

@misc{peng2025sweqa,
      title={{SWE-QA}: Can Language Models Answer Repository-level Code Questions?}, 
      author={Weihan Peng and Yuling Shi and Yuhang Wang and Xinyun Zhang and Beijun Shen and Xiaodong Gu},
      year={2026},
      eprint={2509.14635},
      archivePrefix={arXiv},
      primaryClass={cs.CL},
      url={https://arxiv.org/abs/2509.14635}, 
}

@misc{aa-coding-agents,
 author = {{Artificial Analysis}},
 title = {{{Coding Agent Index}}},
 year = {{{2026}}},
 howpublished = {{\url{https://artificialanalysis.ai/agents/coding-agents/}}}
}

@inproceedings{tirx,
author = {Hou, Bohan and Jin, Hongyi and Wang, Guanjie and Hong, Shushi and Chen, Jinqi and Cai, Yaxing and Yang, Lijie and Ye, Zihao and Ding, Yaoyao and Lai, Ruihang and Mowry, Todd C. and Chen, Tianqi},
title = {{TIRx}: A Unified Tile Primitive ML Compiler Across Heterogeneous Backends},
year = {2027},
isbn = {9798400730122},
publisher = {Association for Computing Machinery},
address = {New York, NY, USA},
url = {https://doi.org/10.1145/3845814.3850059},
doi = {10.1145/3845814.3850059},
booktitle = {Proceedings of the 32nd ACM International Conference on Architectural Support for Programming Languages and Operating Systems, Volume 1},
pages = {955–969},
numpages = {15},
location = {Heraklion, Greece},
series = {ASPLOS '27}
}

@misc{cute-dsl,
  author = {{NVIDIA}},
  title = {{CuTe DSL}: {DSL} Programming Model},
  year = {2026},
  howpublished = {\url{https://docs.nvidia.com/cutlass/latest/media/docs/pythonDSL/cute_dsl.html}}
}

@misc{tirx-harness,
  author = {{MLC Community}},
  title = {{TIRx Harness}: An Open Compiler Harness for Agentic {GPU} Programming},
  year = {2026},
  month = sep,
  howpublished = {\url{https://blog.mlc.ai/2026/09/29/tirx-harness-an-open-compiler-harness-for-agentic-gpu-programming}}
}

@misc{kda,
  author = {{NVIDIA}},
  title = {Kernel Design Agents ({KDA})},
  year = {2026},
  howpublished = {\url{https://github.com/NVlabs/kda}}
}

@misc{slime,
  author       = {Zilin Zhu and Chengxing Xie and Xin Lv and slime Contributors},
  title        = {slime: An {LLM} post-training framework for {RL} Scaling},
  year         = {2025},
  howpublished = {\url{https://github.com/THUDM/slime}},
  note         = {GitHub repository. Corresponding author: Xin Lv},
  urldate      = {2025-06-19}
}

@misc{shao2024deepseekmath,
      title={{DeepSeekMath}: Pushing the Limits of Mathematical Reasoning in Open Language Models}, 
      author={Zhihong Shao and Peiyi Wang and Qihao Zhu and Runxin Xu and Junxiao Song and Xiao Bi and Haowei Zhang and Mingchuan Zhang and Y. K. Li and Y. Wu and Daya Guo},
      year={2024},
      eprint={2402.03300},
      archivePrefix={arXiv},
      primaryClass={cs.CL},
      url={https://arxiv.org/abs/2402.03300}, 
}

@inproceedings{
yu2025dapo,
title={{DAPO}: An Open-Source {LLM} Reinforcement Learning System at Scale},
author={Qiying Yu and Zheng Zhang and Ruofei Zhu and Yufeng Yuan and Xiaochen Zuo and YuYue and Weinan Dai and Tiantian Fan and Gaohong Liu and Juncai Liu and LingJun Liu and Xin Liu and Haibin Lin and Zhiqi Lin and Bole Ma and Guangming Sheng and Yuxuan Tong and Chi Zhang and Mofan Zhang and Ru Zhang and Wang Zhang and Hang Zhu and Jinhua Zhu and Jiaze Chen and Jiangjie Chen and Chengyi Wang and Hongli Yu and Yuxuan Song and Xiangpeng Wei and Hao Zhou and Jingjing Liu and Wei-Ying Ma and Ya-Qin Zhang and Lin Yan and Yonghui Wu and Mingxuan Wang},
booktitle={The Thirty-ninth Annual Conference on Neural Information Processing Systems},
year={2025},
url={https://openreview.net/forum?id=2a36EMSSTp}
}

@inproceedings{
zheng2024sglang,
title={{SGL}ang: Efficient Execution of Structured Language Model Programs},
author={Lianmin Zheng and Liangsheng Yin and Zhiqiang Xie and Chuyue Sun and Jeff Huang and Cody Hao Yu and Shiyi Cao and Christos Kozyrakis and Ion Stoica and Joseph E. Gonzalez and Clark Barrett and Ying Sheng},
booktitle={The Thirty-eighth Annual Conference on Neural Information Processing Systems},
year={2024},
url={https://openreview.net/forum?id=VqkAKQibpq}
}

@misc{liu2025moeparallelfolding,
      title={{MoE} Parallel Folding: Heterogeneous Parallelism Mappings for Efficient Large-Scale {MoE} Model Training with Megatron Core}, 
      author={Dennis Liu and Zijie Yan and Xin Yao and Tong Liu and Vijay Korthikanti and Evan Wu and Shiqing Fan and Gao Deng and Hongxiao Bai and Jianbin Chang and Ashwath Aithal and Michael Andersch and Mohammad Shoeybi and Jiajie Yao and Chandler Zhou and David Wu and Xipeng Li and June Yang},
      year={2026},
      eprint={2504.14960},
      archivePrefix={arXiv},
      primaryClass={cs.LG},
      url={https://arxiv.org/abs/2504.14960}, 
}
\bibliographystyle{plain}

\clearpage
\appendix

\section{{Pretraining Correctness}}
\label{sec:appendix:model-quality}

This appendix validates the {pretraining} correctness of \sys against
\megatron at matched configuration.
We report two complementary measurements: pretraining loss
trajectories (\S\ref{sec:appendix:loss-curve}) and downstream task
accuracy (\S\ref{sec:appendix:downstream}).

\subsection{Pretraining Loss}
\label{sec:appendix:loss-curve}

We pretrain Qwen3-30B-A3B with \megatron and \sys under identical
configuration. \autoref{fig:appendix-loss-curve} reports the full
training configuration alongside the two loss curves; the
trajectories align across the full run, with \megatron showing two
transient spikes that recover within a few steps.

\begin{figure}[t]
  \centering
  \begin{minipage}[c]{0.36\linewidth}
    \centering
    \footnotesize
    \setlength{\tabcolsep}{4pt}
    \begin{tabular}{ll}
      \toprule
      Hardware       & 4$\times$8-H100 \\
      Parallelism    & PP4, DP1, CP1, EP8 \\
      Dataset        & DCLM~\cite{li2024dclm} \\
      Sequence       & 2048 \\
      Precision      & BF16 \\
      Global Batch   & 1024 \\
      Training Steps & 4096 \\
      Warmup Steps   & 128 \\
      Optimizer      & Adam~\cite{kingma2015adam} \\
      Max LR         & $2\times10^{-4}$ \\
      Min LR         & $1\times10^{-5}$ \\
      LR Schedule    & Cosine Decay \\
      Aux Loss       & Global-Level~\cite{qiu2025demonsdetailimplementingload} \\
      Aux Coef       & $1\times10^{-3}$ \\
      \bottomrule
    \end{tabular}
  \end{minipage}
  \hfill
  \begin{minipage}[c]{0.62\linewidth}
    \centering
    \includegraphics[width=\linewidth]{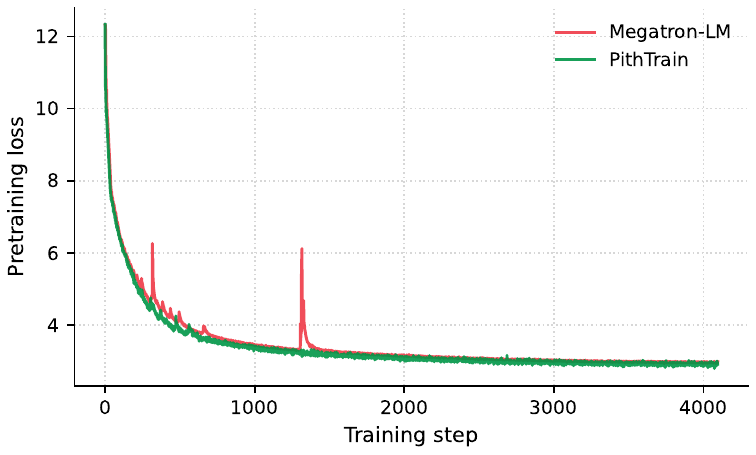}
  \end{minipage}
  \caption{Training configuration (left) and pretraining loss curves
  (right) for Qwen3-30B-A3B trained with \megatron and \sys.}
  \label{fig:appendix-loss-curve}
\end{figure}

\subsection{Downstream Accuracy}
\label{sec:appendix:downstream}

We evaluate downstream task accuracy across six standard
benchmarks:
OpenBookQA~\cite{mihaylov2018openbookqa} and
WinoGrande~\cite{sakaguchi2020winogrande} in the 0-shot setting, and
ARC-Challenge, ARC-Easy~\cite{clark2018arc},
HellaSwag~\cite{zellers2019hellaswag}, and PIQA~\cite{bisk2020piqa} in
the 10-shot setting. \autoref{fig:appendix-downstream} plots
accuracy for each task; \megatron and \sys achieve comparable
accuracy within statistical noise across all six benchmarks at every
checkpoint.

\begin{figure}[t]
  \centering
  \begin{subfigure}{0.32\linewidth}\includegraphics[width=\linewidth]{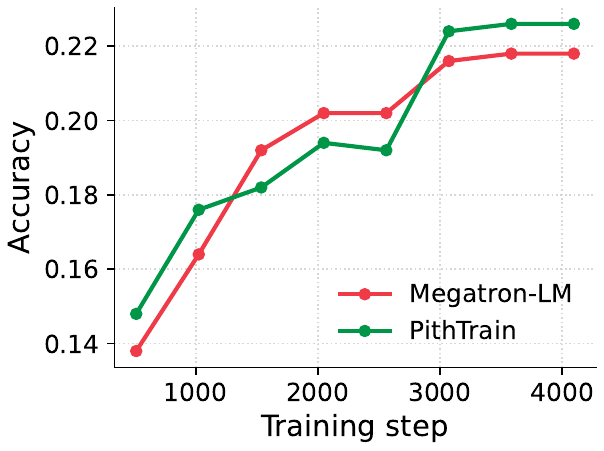}\caption{OpenBookQA (0-shot)}\end{subfigure}
  \hfill
  \begin{subfigure}{0.32\linewidth}\includegraphics[width=\linewidth]{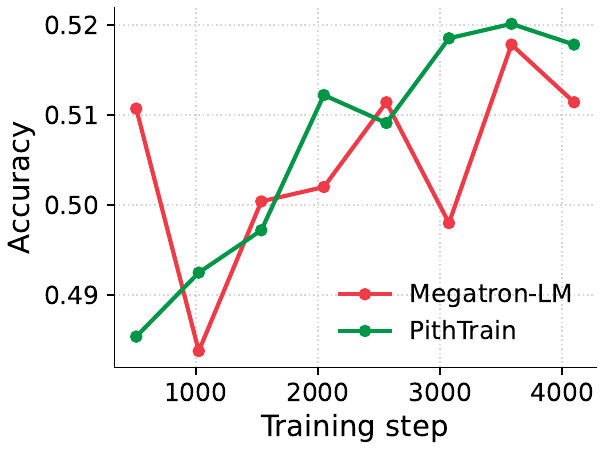}\caption{WinoGrande (0-shot)}\end{subfigure}
  \hfill
  \begin{subfigure}{0.32\linewidth}\includegraphics[width=\linewidth]{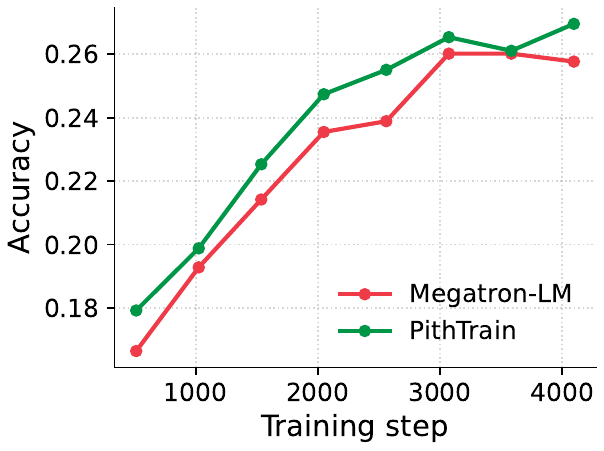}\caption{ARC-Challenge (10-shot)}\end{subfigure}

  \medskip
  \begin{subfigure}{0.32\linewidth}\includegraphics[width=\linewidth]{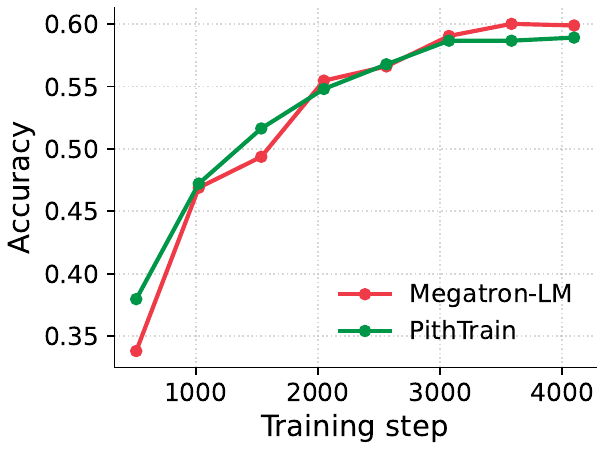}\caption{ARC-Easy (10-shot)}\end{subfigure}
  \hfill
  \begin{subfigure}{0.32\linewidth}\includegraphics[width=\linewidth]{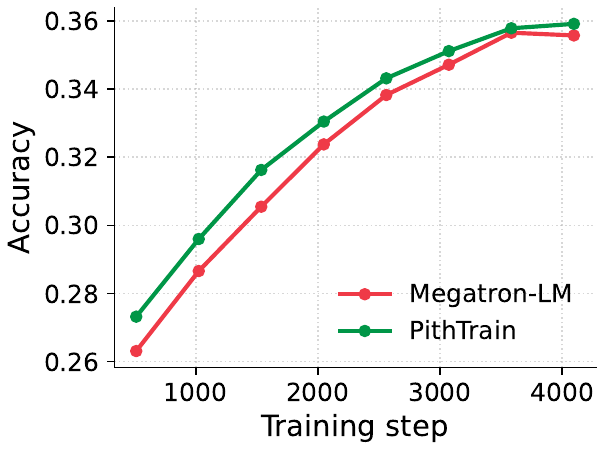}\caption{HellaSwag (10-shot)}\end{subfigure}
  \hfill
  \begin{subfigure}{0.32\linewidth}\includegraphics[width=\linewidth]{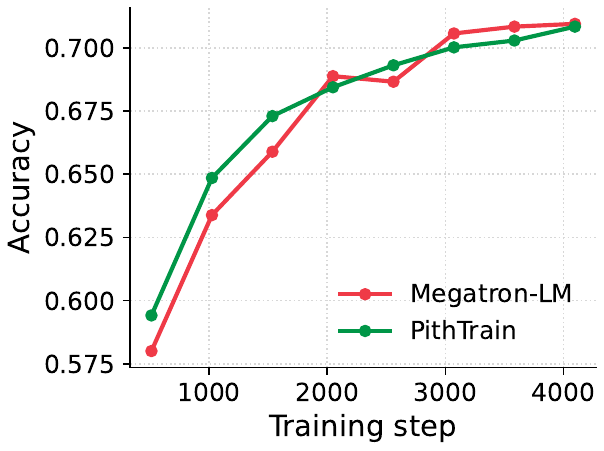}\caption{PIQA (10-shot)}\end{subfigure}

  \caption{Downstream task accuracy for Qwen3-30B-A3B trained with
  \megatron and \sys.}
  \label{fig:appendix-downstream}
\end{figure}

\clearpage
\par\begingroup
\section{Reinforcement Learning}
\label{sec:appendix:rl}

We integrate \sys as a training backend in slime~\cite{slime}, which
pairs a trainer with SGLang~\cite{zheng2024sglang} rollout engines,
and run asynchronous GRPO~\cite{shao2024deepseekmath} on
Qwen3-30B-A3B-Base with \megatron and \sys under identical
configuration (\autoref{fig:appendix-rl}a). Rollouts are
generated on a separate H200 node, and the trainer pushes updated weights
to the engines after every step.

\MyPara{Identical data.} We first record 32 asynchronous batches and
replay them through both trainers, so that both optimize the same
objective on the same data. Across the 32 steps, the per-step policy
entropy, policy-gradient loss, gradient norm, clip fraction, and KL
divergence show no growing discrepancy between backends.
Policy entropy differs by about 0.05\% between backends, compared
with about 0.03\% between two \sys launches.

\MyPara{End-to-end training.} We then train each backend
independently for 512 steps. \autoref{fig:appendix-rl}b--c show
that training reward and AIME 2024 accuracy track each other
throughout; the mean reward over the run is 0.454 for \sys and 0.447
for \megatron. On a paired evaluation of the final checkpoints (30
AIME 2024 problems, 64 samples each), \sys and \megatron reach 0.226
and 0.231 against 0.104 for the base model. The difference,
$-0.005 \pm 0.007$ (standard error; 95\% CI $[-0.020, 0.008]$), is not significant,
while both backends improve on the base model by about 12--13 percentage points.

\begin{figure}[t]
  \centering
  \begin{subfigure}[b]{0.32\linewidth}
    \centering
    \scriptsize
    \setlength{\tabcolsep}{3pt}
    \renewcommand{\arraystretch}{1.05}%
    \begin{tabular}{ll}
      \toprule
      Hardware       & 2$\times$8-H200 \\
      Parallelism    & PP2, DP4, EP4 \\
      Dataset        & DAPO-Math-17K~\cite{yu2025dapo} \\
      Algorithm      & Async GRPO \\
      Batch          & 64 prompts $\times$ 8 \\
      Max Response   & 8192 \\
      Training Steps & 512 \\
      Optimizer      & AdamW \\
      LR             & $1\times10^{-6}$ \\
      PPO Clip       & 0.2 / 0.28 \\
      \bottomrule
    \end{tabular}
    \vspace{6pt}
    \caption{Configuration}
  \end{subfigure}
  \hfill
  \begin{subfigure}[b]{0.32\linewidth}\includegraphics[width=\linewidth]{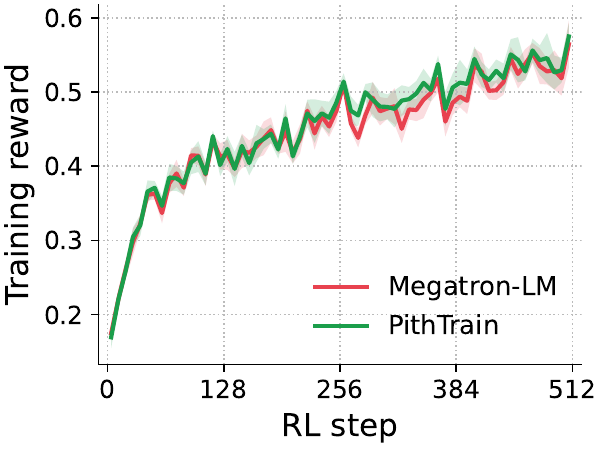}\caption{Training reward}\end{subfigure}
  \hfill
  \begin{subfigure}[b]{0.32\linewidth}\includegraphics[width=\linewidth]{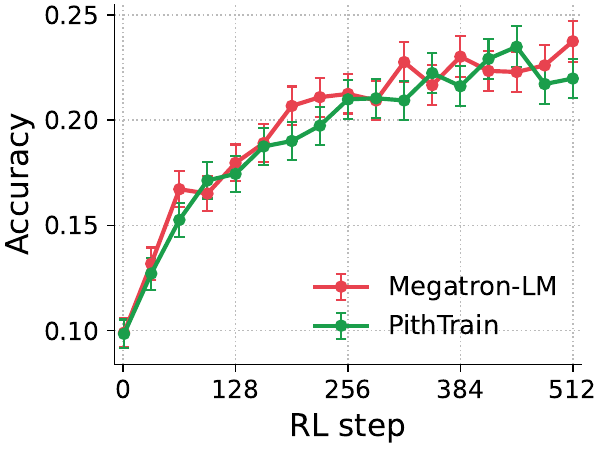}\caption{AIME 2024}\end{subfigure}
  \caption{Asynchronous GRPO on Qwen3-30B-A3B-Base with \megatron and
  \sys. Training reward is an 8-step mean $\pm$ s.e.; AIME 2024 uses
  30 problems $\times$ 64 samples.}
  \label{fig:appendix-rl}
\end{figure}
\par\endgroup

\section{Task Suite}
\label{sec:appendix:tasks}

This appendix expands the per-category task descriptions and the
correctness checks used to validate each attempt. All
operate-and-profile and new-feature tasks share a fixed training
configuration: the base model is
DeepSeek-V2-Lite~\cite{deepseekai2024deepseekv2strongeconomicalefficient}, the
parallelism mesh is PP=4, EP=2, DP=1, sequence length 2048, global
batch size 1024, precision BF16.

\subsection{Q\&A Tasks}
\label{sec:appendix-codebaseqa}

Each Q\&A task asks the agent to locate where a specific behavior
lives in the framework codebase. The agent receives a single
prompt consisting of the \emph{universal instruction} below followed
verbatim by the full query text for that task (Q1--Q12 in
\S\ref{sec:appendix:tasks:qa-descriptions}), and has read-only access
to \texttt{Read}, \texttt{Grep}, \texttt{Glob}, and \texttt{Bash};
tools that modify the working tree (\texttt{Edit}, \texttt{Write},
\texttt{NotebookEdit}) are explicitly disabled. Correctness is
validated by two independent human graders
(\S\ref{sec:appendix:tasks:qa-correctness}).

\MyPara{Universal instruction (prepended to every query).}
Your task is to explore this training framework codebase and answer
the question that follows. Every claim must be verified against the
code on disk before you state it: use your CLI tools (Read, Grep,
Glob, Bash) to locate the exact symbol, file, and line, then cite the
\texttt{path/to/file.py:LINE} you actually read. Wrap your final
response in \texttt{<final\_answer>} tags, trace the execution path
step by step, and give one citation per claim. If a feature is
genuinely absent from this codebase, say so explicitly and cite
negative evidence (the grep command and pattern that returned zero
matches, or the file you inspected that does not contain it). A
correct ``absent, verified'' answer is preferred over a fabricated
one.

\subsubsection{Task Descriptions}
\label{sec:appendix:tasks:qa-descriptions}

\MyPara{Q1: Process Groups / Device Mesh.}
Trace the sequence of function calls from the main training entry
script down to the initialization of the PyTorch distributed process
groups or DeviceMesh. Detail the exact file paths, function names,
and line numbers where the world size and ranks for the parallel
groups present in this codebase (any of DP, TP, PP, EP, CP that this
repo actually supports) are assigned.

\MyPara{Q2: Configuration Propagation.}
Locate the exact file and line numbers where the user-provided
configuration for \texttt{hidden\_size} or \texttt{dim} is parsed.
Trace how this specific variable propagates down to the instantiation
of the first Transformer block.

\MyPara{Q3: Data Loading \& Sharding.}
Trace the initialization of the dataset and dataloader. Identify the
file, class, and line number where the global dataset is sliced or
partitioned among the Data Parallel ranks to ensure each rank
receives a unique, non-overlapping subset of data.

\MyPara{Q4: Distributed Seed Management.}
Find where the random seeds are set for data loading versus model
initialization. Provide the exact file paths and line numbers, and
trace how seeds are differentiated across different parallel groups
(or, if they are deliberately shared, where and why).

\MyPara{Q5: Attention Kernel Dispatch.}
Locate the core attention module. Trace the logic that selects
between an optimised kernel (FlashAttention, PyTorch SDPA flash
backend, ring attention, etc.) and any fallback math implementation.
Identify the exact file, class, line number, and configuration
flag/condition controlling this dispatch. If the dispatch is
delegated to the PyTorch \texttt{scaled\_dot\_product\_attention}, say
so and cite the call site.

\MyPara{Q6: RoPE Implementation.}
Find where the Rotary Positional Embedding (RoPE) is implemented.
Locate the function that precomputes the frequency tensor
(sine/cosine cache). Provide the file path, class/function name, and
exact line numbers of the tensor operations where the rotary
transform is applied to the Query and Key tensors.

\MyPara{Q7: SwiGLU / MLP Block.}
Locate the Feed-Forward Network (FFN) implementation. Identify the
file and line number where the SwiGLU activation (or equivalent gated
linear unit) is mathematically applied. Trace how the up-projection
tensor is chunked or split before the activation function.

\MyPara{Q8: Normalization Placement.}
Find the implementation of the main Transformer layer. Provide the
file and exact line numbers where the normalization (RMSNorm or
LayerNorm) is applied before the attention module. Detail where the
\texttt{epsilon} (eps) term is added for numerical stability.

\MyPara{Q9: Context / Sequence Parallelism.}
Find where sequence parallelism (or context parallelism) is handled.
Identify the exact file and line numbers where the sequence dimension
of the input tensor is sharded, gathered, or scattered across ranks
during the forward and backward passes. If the codebase does not
support CP/SP, say so and cite the negative evidence.

\MyPara{Q10: FSDP / DDP Wrapping.}
Locate the exact file and line numbers where the core model is
wrapped with Fully Sharded Data Parallel (FSDP1 or FSDP2) or
Distributed Data Parallel (DDP) wrappers, and identify the
device-mesh axes used for sharding/replication.

\MyPara{Q11: Global Gradient Clipping.}
Trace the logic for global gradient norm clipping. Find the file,
function, and exact line number where the global norm of the
gradients is computed (including any reduction across distributed
ranks) before the optimizer step.

\MyPara{Q12: Distributed Checkpoint Serialization.}
Locate the model checkpoint saving logic. Detail whether the system
uses a rank-0 gather approach or distributed sharded saving (e.g.,
PyTorch DCP). Provide the exact file path and line numbers where the
disk serialization occurs.

\subsubsection{Correctness Checks}
\label{sec:appendix:tasks:qa-correctness}

Each Q\&A answer the agent returns cites the file paths, function
names, and line numbers in the framework codebase that implement
the behavior the question asks about. Two human graders
independently trace these citations: they open each cited code
location and verify that the code there does what the agent claims
in the answer. An attempt is recorded as \emph{satisfied} when both
graders confirm every citation against the code on disk;
disagreements are resolved by a third reader who looks only at the
cited evidence. All 108 attempts (12 questions $\times$ 3
frameworks $\times$ 3 attempts) were judged satisfied by both
graders.

\subsection{Operate-and-profile Tasks}
\label{sec:appendix:tasks:operate}

Each operate-and-profile task asks the agent to drive a real
training-system workflow end-to-end, so success requires the agent
to set up the environment, run the framework as intended, and
produce the expected artifact. The four tasks span the workflows
a researcher typically performs before any code change: getting
the framework running, executing a full train-and-evaluate
pipeline, instrumenting the model to capture behavior, and
profiling the system to surface bottlenecks.

\subsubsection{Task Descriptions}
\label{sec:appendix:tasks:operate-descriptions}

\MyPara{Getting Started.}
Set up the Python environment for the framework and run the
provided 5-step smoke training script. Success means the script
reaches step 5 with a finite loss. The agent must install all
dependencies the script needs so that running \texttt{bash
train.sh} as-is succeeds; \texttt{train.sh} itself documents best
practices for training MoE models and is read-only. Pre-tokenized
DCLM and the converted base-model checkpoint are pre-staged.

\MyPara{Train and Evaluate.}
Drive the full setup-train-export-evaluate pipeline for
the base model. The agent trains from random initialization for
25 steps, exports the resulting checkpoint to HuggingFace format,
and runs lm-evaluation-harness HellaSwag (zero-shot) on it via
vLLM. The task tests pipeline correctness, not model quality: the
HellaSwag score is expected to be near-random because 25 steps from
random init produces a barely-trained model, and the agent must
produce \emph{some} score from a working pipeline. Initialization
must be random; everything outside the fixed mesh and step count
(LR, optimizer, scheduler, data preprocessing) is left to the
agent.

\MyPara{Collect Routing Trace.}
Instrument training to dump the per-token MoE routing trace for
the first 8M training tokens. The routing
decision in each MoE layer,
the top-$k$ expert IDs and their gating weights, must be captured
for every token in the global batch. Training resumes from the
released HuggingFace checkpoint so the router is in its trained,
load-balanced regime; routing decisions are model-intrinsic and
valid from step 1, so no warmup is required. The output schema is
one \texttt{step-<step\_id:08d>.npz} file per step under
\texttt{workspace/<framework>/routing-traces/}, each containing the
expert-ID and gating-weight arrays.

\MyPara{Report Heavy Kernels.}
Profile a 7-step training run with Nsight Systems
and identify the top 3 most expensive CUDA kernels by total GPU
time, aggregated across all 8 ranks. Training resumes from the
released HuggingFace checkpoint; the agent profiles only step 7
because steps 1--6 are warmup for cudagraph capture, NCCL
handshake, and allocator priming. The output is a single CSV at
\texttt{workspace/<framework>/heavy-kernels/top-kernels.csv} with
header \texttt{kernel\_name,total\_time\_ms,instances,mean\_time\_ms}
and exactly three rows sorted by \texttt{total\_time\_ms}
descending, plus the raw \texttt{profile.nsys-rep} so the result is
reproducible.

\subsubsection{Correctness Checks}
\label{sec:appendix:tasks:operate-correctness}

Each operate-and-profile task verifies the artifact the agent
produces, not the path the agent took to produce it. We use a
mix of programmatic checks baked into the task harness and human
inspection where the artifact is non-textual.

\MyPara{Getting Started.}
The task succeeds when running \texttt{train.sh} reaches step 5
with a finite loss. After the agent finishes installing
dependencies, the harness re-runs the (read-only) training script
and parses the resulting log; the artifact is accepted only if
step 5 prints a finite loss value.

\MyPara{Train and Evaluate.}
The harness ships a read-only \texttt{evaluate.sh} script that
consumes the pipeline output produced by the agent: the agent
trains for 25
steps from random init, exports the resulting checkpoint to
HuggingFace format, and \texttt{evaluate.sh} loads the export into
vLLM and runs lm-evaluation-harness HellaSwag (zero-shot). An
attempt is satisfied when \texttt{evaluate.sh} runs to completion
and writes a finite HellaSwag score; the score is not required to
clear any quality threshold, since 25 steps from random
initialization is expected to yield near-random accuracy.

\MyPara{Collect Routing Trace.}
The harness verifies that four
\texttt{step-<step\_id:08d>.npz} files are present at
\texttt{workspace/<framework>/routing-traces/} and that each
carries expert-ID and gating-weight arrays of the correct shape
for every MoE layer over every token in the global batch. A
programmatic check additionally confirms that the expert-ID
values are within the valid range \texttt{[0, num\_experts)} for
the MoE configuration of the model, and that the gating weights
are non-negative and sum to 1 over the top-$k$ selected experts
for every token.

\MyPara{Report Heavy Kernels.}
The agent submits both \texttt{top-kernels.csv} and the raw
\texttt{profile.nsys-rep}. The harness checks the CSV
programmatically against the prescribed schema (header, exactly
three rows sorted by total time descending). The kernel names
themselves are validated by a human reader who opens
\texttt{profile.nsys-rep} in the Nsight Systems profiler GUI,
reads the CUDA GPU Kernel Summary in the Stats System view, and
confirms that the top three kernel names in the CSV produced by
the agent match the top three entries shown by the profiler.

All 36 attempts (4 tasks $\times$ 3 frameworks $\times$ 3
attempts) produced an artifact accepted by both the harness and
the human reader.

\subsection{New-feature Tasks}
\label{sec:appendix:tasks:new-feature}

Each new-feature task mimics the workflow of a researcher or ML
engineer integrating a recently published modeling architecture
into a training framework. The agent is given the same materials
such an engineer would normally consult: the arXiv paper and the
reference implementation. From these inputs, the agent must
produce a training script that integrates the new feature into the
base model and runs on DCLM~\cite{li2024dclm} under the fixed
configuration above.

The four tasks are chosen for both coverage and a fair
starting point. Two revise the attention mechanism and two require
changes at the FFN (MoE) layer, so the suite spans the architectural
subsystems a training framework must accommodate. None of the four
architectures had been integrated into any of the three frameworks
we compare at the time of this study, so all frameworks start the
task from the same point and the comparison reflects only how the
design of each framework affects integration effort.

Correctness is validated on two axes. First, the cross-entropy loss
must decrease across the 64-step run and remain finite (no explosion,
no NaN), evidence that the modified training pipeline produces a
learnable model. Second, the changes made by the
agent must satisfy three task-specific rules, each validating one
required component of the new feature
(\S\ref{sec:appendix:tasks:rules}); the rules target the
architectural elements that distinguish each new feature from the
baseline framework, ensuring the agent has implemented the intended
mechanism rather than producing a passing-loss reading from an
unchanged model. An attempt is recorded as \emph{satisfied} when both
axes hold.

\subsubsection{Task Descriptions}
\label{sec:appendix:tasks:descriptions}

\MyPara{Diff~\cite{ye2024differential}.}
Differential attention replaces a single softmax-attention map with
the difference of two parallel softmax maps,
$\mathrm{Attn}(Q,K,V) = (\mathrm{softmax}(Q_1 K_1^\top) - \lambda\,
\mathrm{softmax}(Q_2 K_2^\top))\,V$, where $\lambda$ is a learned
per-head scalar. The construction cancels common-mode attention
noise, sharpening which tokens receive mass. Integrating it
involves splitting the per-head $Q$/$K$ projections in half and
introducing $\lambda$ as a trainable parameter with the published
initialisation schedule.

\MyPara{DynMoE~\cite{guo2024dynamic}.}
Standard MoE fixes the number of activated experts $k$ per token.
DynMoE replaces top-$k$ selection with a per-expert sigmoid gate and
a learned threshold, so the count of active experts varies per
token. Integrating it involves replacing the top-$k$ routing
kernel of the framework and reformulating the load-balancing
auxiliary loss for a variable-$k$ regime.

\MyPara{MoBA~\cite{lu2025moba}.}
MoBA partitions the key/value sequence into fixed-size blocks
and routes each query to the top-$k$ blocks
selected by {query--block affinity scores}, yielding sub-quadratic attention
cost in the sequence length. Integrating it involves
inserting block-level routing between the $Q$/$K$ projection and
the attention computation, composing with the existing attention
backend in the framework, and preserving
causal masking inside each selected block.

\MyPara{MoE++~\cite{jin2025moeplusplus}.}
MoE++ augments a standard MoE expert pool with three
zero-computation expert types (\emph{zero}, \emph{copy}, and
\emph{constant}), allowing easy tokens to be routed past the
feed-forward layer entirely. Integrating it involves extending the
expert pool definition with the zero-computation variants, widening
the router output to cover them, and adjusting the load-balancing
loss to prevent the zero experts from absorbing all easy tokens.

\subsubsection{Rule-based Correctness Checks}
\label{sec:appendix:tasks:rules}

Each new-feature task decomposes into three required components,
the architectural elements that distinguish the new feature from
the baseline framework. The rules below name each component and
describe how it should be implemented; they were fixed before
inspecting any attempt. Each
attempt is judged by an independent
\texttt{claude-opus-4-7} session at \texttt{xhigh} effort, given
the three rules and the git diff produced by the agent against the
main branch of the framework; the judge returns PASS or FAIL per
rule with a one-sentence justification quoting a specific code
reference, and an attempt is recorded as \emph{satisfied} when all
three rules pass. We additionally inspect every verdict and its
cited justification by hand: a human reader opens the diff at the
cited file and confirms the line evidence for both PASS judgements
and any FAIL or partial verdicts. Across all 36 attempts
(4 tasks $\times$ 3 frameworks $\times$ 3 attempts), every attempt
satisfies all three rules of its task.

\MyPara{Diff.}
\begin{itemize}
  \setlength\itemsep{1pt}
  \item[R1.] The attention forward path produces two separate softmax outputs that are combined as $\mathrm{attn}_1 - \lambda \cdot \mathrm{attn}_2$ (literal subtraction with a learnable coefficient).
  \item[R2.] A learnable parameter ($\lambda$, e.g.\ named \texttt{lambda} or \texttt{lambda\_init}) is registered as an \texttt{nn.Parameter} with shape compatible with per-head broadcasting.
  \item[R3.] The $Q$/$K$ projections are split into two halves, either via an output dimension of $2 \cdot n_\mathrm{heads} \cdot d_\mathrm{head}$ followed by a split, or via two separate projection layers.
\end{itemize}

\MyPara{DynMoE.}
\begin{itemize}
  \setlength\itemsep{1pt}
  \item[R1.] The router uses per-expert sigmoid gates (or sigmoid-gated activation), not pure softmax with top-$k$ selection.
  \item[R2.] A learnable threshold parameter for expert activation is registered as an \texttt{nn.Parameter} (often named \texttt{tau}, \texttt{threshold}, or \texttt{gate\_threshold}).
  \item[R3.] Active-expert selection is not a fixed \texttt{top\_k = N}; the count depends on which experts pass the threshold (boolean mask or variable-length selection).
\end{itemize}

\MyPara{MoBA.}
\begin{itemize}
  \setlength\itemsep{1pt}
  \item[R1.] The $K$/$V$ sequence is partitioned into fixed-size blocks (a \texttt{reshape} or \texttt{view} into shape \texttt{[\dots, $n_\mathrm{blocks}$, $B$, \dots]} or equivalent).
  \item[R2.] Each query computes a per-block score (typically $\mathrm{query} \cdot \mathrm{pooled\_block\_key}$) followed by top-$k$ block selection.
  \item[R3.] Final attention is computed only over the selected blocks, with causal masking preserved at block boundaries (each query attends only to blocks at positions $\leq$ its own).
\end{itemize}

\MyPara{MoE++.}
\begin{itemize}
  \setlength\itemsep{1pt}
  \item[R1.] The expert pool includes at least one zero-computation expert type (\emph{zero}, \emph{copy}, or \emph{constant}), identifiable by class name, string literal, or a specialised expert factory.
  \item[R2.] The router emits logits over a pool that contains both regular FFN experts and zero-computation experts, so the router learns to dispatch tokens to the no-op variants. The zero-computation experts may be added on top of the existing pool (router output dim grows to $n_{\mathrm{FFN}} + n_{\mathrm{zero}}$) or carved out of an unchanged total expert count.
  \item[R3.] The load-balancing auxiliary loss handles zero experts explicitly: they are excluded from the balance term, weighted differently, or balanced under a separate penalty.
\end{itemize}

\section{Per-Task Results}
\label{sec:appendix:per-task-results}

\subsection{{Claude Code}}
\label{sec:appendix:claude}

{This subsection reports per-attempt agent effort for Claude Code
across all 20 tasks.} Tables~\ref{tab:appendix-per-attempt-codebaseqa}
and~\ref{tab:appendix-per-attempt-codebaseqa-2} cover the 12 Q\&A
tasks (split 6/6 across two pages),
\autoref{tab:appendix-per-attempt-operate} covers the four
operate-and-profile tasks, and
\autoref{tab:appendix-per-attempt-new-feature} covers the four
new-feature tasks. Each row is one independent attempt;
per-task medians appear in the corresponding main-text tables in
\S\ref{sec:eval:agent}. Session Duration and Active GPU Time are
reported in minutes. Lower is better on every metric.

\begin{table}[t]
  \centering
  \footnotesize
  \caption{Per-attempt agent effort on the Q\&A tasks Q1--Q6 (\S\ref{sec:eval:agent}). Each row is one independent attempt. Session Duration is in minutes.}
  \label{tab:appendix-per-attempt-codebaseqa}
  \setlength{\tabcolsep}{4pt}
  \begin{tabular}{@{}llr|rrrr@{}}
    \toprule
    Task
      & Framework
      & Attempt
      & \makecell[br]{Session\\Duration}
      & \makecell[br]{Agent\\Turns}
      & \makecell[br]{Per-Turn\\Context}
      & \makecell[br]{Output\\Tokens} \\
    \midrule
    \multirow{9}{*}{Q1: Process Groups / Device Mesh}
      & \megatron     & 1st & 2.0 & 33 & 47.8K & 9.0K \\
      &                               & 2nd & 2.2 & 26 & 45.8K & 9.7K \\
      &                               & 3rd & 2.1 & 34 & 44.5K & 9.0K \\
      & \torchtitan   & 1st & 2.8 & 18 & 44.3K & 7.1K \\
      &                               & 2nd & 3.1 & 19 & 45.2K & 7.0K \\
      &                               & 3rd & 3.0 & 15 & 42.9K & 7.3K \\
      & \sys          & 1st & 1.6 & 15 & 33.4K & 4.1K \\
      &                               & 2nd & 1.4 & 13 & 33.1K & 3.5K \\
      &                               & 3rd & 2.0 & 17 & 35.1K & 5.3K \\
    \midrule
    \multirow{9}{*}{Q2: Configuration Propagation}
      & \megatron     & 1st & 2.9 & 54 & 43.4K & 10.7K \\
      &                               & 2nd & 1.9 & 37 & 37.6K & 8.4K \\
      &                               & 3rd & 3.2 & 54 & 41.2K & 12.0K \\
      & \torchtitan   & 1st & 2.8 & 25 & 42.6K & 6.4K \\
      &                               & 2nd & 3.0 & 34 & 43.5K & 7.8K \\
      &                               & 3rd & 3.5 & 28 & 45.0K & 9.2K \\
      & \sys          & 1st & 2.0 & 18 & 35.7K & 4.5K \\
      &                               & 2nd & 2.9 & 28 & 35.9K & 7.6K \\
      &                               & 3rd & 2.2 & 18 & 36.8K & 5.4K \\
    \midrule
    \multirow{9}{*}{Q3: Data Loading \& Sharding}
      & \megatron     & 1st & 1.0 & 10 & 32.1K & 3.8K \\
      &                               & 2nd & 0.8 & 14 & 31.3K & 3.0K \\
      &                               & 3rd & 0.9 & 14 & 30.8K & 3.4K \\
      & \torchtitan   & 1st & 1.2 & 14 & 35.0K & 3.2K \\
      &                               & 2nd & 1.8 & 21 & 38.2K & 4.0K \\
      &                               & 3rd & 1.2 & 14 & 36.1K & 3.5K \\
      & \sys          & 1st & 1.4 & 13 & 34.5K & 3.5K \\
      &                               & 2nd & 1.3 & 15 & 29.1K & 3.4K \\
      &                               & 3rd & 1.5 & 8 & 34.1K & 2.9K \\
    \midrule
    \multirow{9}{*}{Q4: Distributed Seed Management}
      & \megatron     & 1st & 2.2 & 36 & 37.5K & 8.8K \\
      &                               & 2nd & 2.2 & 37 & 49.4K & 10.0K \\
      &                               & 3rd & 2.5 & 34 & 41.3K & 10.4K \\
      & \torchtitan   & 1st & 3.1 & 22 & 37.9K & 6.5K \\
      &                               & 2nd & 3.3 & 26 & 39.7K & 6.9K \\
      &                               & 3rd & 3.9 & 32 & 41.7K & 7.9K \\
      & \sys          & 1st & 2.2 & 15 & 34.0K & 5.0K \\
      &                               & 2nd & 2.3 & 19 & 31.1K & 5.2K \\
      &                               & 3rd & 2.5 & 16 & 30.7K & 5.6K \\
    \midrule
    \multirow{9}{*}{Q5: Attention Kernel Dispatch}
      & \megatron     & 1st & 3.7 & 63 & 56.0K & 14.0K \\
      &                               & 2nd & 2.8 & 48 & 46.5K & 10.7K \\
      &                               & 3rd & 1.7 & 27 & 39.8K & 6.6K \\
      & \torchtitan   & 1st & 2.1 & 20 & 41.7K & 4.8K \\
      &                               & 2nd & 2.0 & 19 & 43.7K & 5.1K \\
      &                               & 3rd & 2.0 & 15 & 41.2K & 4.7K \\
      & \sys          & 1st & 1.9 & 20 & 34.2K & 4.4K \\
      &                               & 2nd & 1.9 & 24 & 33.7K & 4.6K \\
      &                               & 3rd & 2.0 & 21 & 33.8K & 5.0K \\
    \midrule
    \multirow{9}{*}{Q6: RoPE Implementation}
      & \megatron     & 1st & 1.1 & 12 & 36.4K & 4.5K \\
      &                               & 2nd & 1.1 & 12 & 35.9K & 4.5K \\
      &                               & 3rd & 0.9 & 12 & 36.4K & 3.2K \\
      & \torchtitan   & 1st & 0.7 & 4 & 29.9K & 1.9K \\
      &                               & 2nd & 1.0 & 4 & 29.9K & 2.6K \\
      &                               & 3rd & 1.0 & 4 & 29.9K & 2.3K \\
      & \sys          & 1st & 1.2 & 8 & 26.7K & 2.6K \\
      &                               & 2nd & 1.3 & 11 & 27.9K & 3.3K \\
      &                               & 3rd & 2.0 & 18 & 31.8K & 5.6K \\
    \bottomrule
  \end{tabular}
\end{table}

\begin{table}[t]
  \centering
  \footnotesize
  \caption{Per-attempt agent effort on the Q\&A tasks (continued, Q7--Q12).}
  \label{tab:appendix-per-attempt-codebaseqa-2}
  \setlength{\tabcolsep}{4pt}
  \begin{tabular}{@{}llr|rrrr@{}}
    \toprule
    Task
      & Framework
      & Attempt
      & \makecell[br]{Session\\Duration}
      & \makecell[br]{Agent\\Turns}
      & \makecell[br]{Per-Turn\\Context}
      & \makecell[br]{Output\\Tokens} \\
    \midrule
    \multirow{9}{*}{Q7: SwiGLU / MLP Block}
      & \megatron     & 1st & 1.0 & 10 & 31.5K & 3.2K \\
      &                               & 2nd & 0.7 & 8 & 31.4K & 2.5K \\
      &                               & 3rd & 0.7 & 9 & 34.1K & 2.6K \\
      & \torchtitan   & 1st & 1.6 & 11 & 30.0K & 3.2K \\
      &                               & 2nd & 1.4 & 13 & 30.2K & 3.8K \\
      &                               & 3rd & 0.9 & 12 & 30.0K & 3.4K \\
      & \sys          & 1st & 2.1 & 15 & 31.5K & 4.6K \\
      &                               & 2nd & 1.6 & 13 & 40.3K & 4.2K \\
      &                               & 3rd & 1.5 & 14 & 30.1K & 3.8K \\
    \midrule
    \multirow{9}{*}{Q8: Normalization Placement}
      & \megatron     & 1st & 1.0 & 15 & 30.3K & 4.1K \\
      &                               & 2nd & 2.6 & 20 & 30.8K & 4.7K \\
      &                               & 3rd & 1.5 & 25 & 34.2K & 5.5K \\
      & \torchtitan   & 1st & 1.0 & 11 & 32.7K & 3.5K \\
      &                               & 2nd & 0.7 & 8 & 30.3K & 2.5K \\
      &                               & 3rd & 1.1 & 14 & 33.7K & 3.5K \\
      & \sys          & 1st & 1.1 & 9 & 27.7K & 3.0K \\
      &                               & 2nd & 1.3 & 9 & 28.7K & 3.2K \\
      &                               & 3rd & 1.4 & 11 & 29.9K & 3.6K \\
    \midrule
    \multirow{9}{*}{Q9: Context / Sequence Parallelism}
      & \megatron     & 1st & 4.7 & 60 & 45.6K & 6.8K \\
      &                               & 2nd & 3.1 & 34 & 45.2K & 7.6K \\
      &                               & 3rd & 3.3 & 29 & 44.6K & 7.8K \\
      & \torchtitan   & 1st & 1.5 & 20 & 38.6K & 5.8K \\
      &                               & 2nd & 2.0 & 16 & 38.6K & 7.5K \\
      &                               & 3rd & 1.6 & 16 & 37.1K & 6.0K \\
      & \sys          & 1st & 2.9 & 22 & 37.6K & 7.1K \\
      &                               & 2nd & 2.2 & 16 & 37.0K & 5.1K \\
      &                               & 3rd & 1.7 & 16 & 35.1K & 4.1K \\
    \midrule
    \multirow{9}{*}{Q10: FSDP / DDP Wrapping}
      & \megatron     & 1st & 2.8 & 19 & 37.8K & 7.0K \\
      &                               & 2nd & 2.3 & 21 & 38.6K & 5.6K \\
      &                               & 3rd & 2.7 & 16 & 35.5K & 5.9K \\
      & \torchtitan   & 1st & 1.6 & 24 & 44.1K & 6.0K \\
      &                               & 2nd & 1.3 & 17 & 36.7K & 4.9K \\
      &                               & 3rd & 1.5 & 16 & 38.8K & 5.1K \\
      & \sys          & 1st & 1.1 & 9 & 26.9K & 2.6K \\
      &                               & 2nd & 1.2 & 12 & 30.2K & 2.9K \\
      &                               & 3rd & 1.4 & 11 & 29.4K & 3.0K \\
    \midrule
    \multirow{9}{*}{Q11: Global Gradient Clipping}
      & \megatron     & 1st & 1.4 & 10 & 31.9K & 3.2K \\
      &                               & 2nd & 1.2 & 12 & 31.3K & 3.1K \\
      &                               & 3rd & 1.0 & 10 & 31.0K & 2.9K \\
      & \torchtitan   & 1st & 0.7 & 9 & 30.0K & 2.5K \\
      &                               & 2nd & 0.7 & 8 & 30.2K & 2.7K \\
      &                               & 3rd & 0.7 & 7 & 30.1K & 2.5K \\
      & \sys          & 1st & 0.8 & 6 & 25.6K & 2.0K \\
      &                               & 2nd & 0.8 & 7 & 25.8K & 1.9K \\
      &                               & 3rd & 0.8 & 6 & 25.5K & 2.0K \\
    \midrule
    \multirow{9}{*}{Q12: Distributed Checkpoint Serialization}
      & \megatron     & 1st & 2.0 & 19 & 37.2K & 5.1K \\
      &                               & 2nd & 1.8 & 18 & 36.2K & 4.5K \\
      &                               & 3rd & 2.8 & 17 & 37.1K & 6.4K \\
      & \torchtitan   & 1st & 0.6 & 5 & 33.9K & 2.2K \\
      &                               & 2nd & 0.8 & 5 & 33.9K & 2.6K \\
      &                               & 3rd & 0.6 & 5 & 33.9K & 2.2K \\
      & \sys          & 1st & 1.2 & 10 & 31.1K & 2.5K \\
      &                               & 2nd & 1.1 & 9 & 30.0K & 2.6K \\
      &                               & 3rd & 1.4 & 10 & 35.3K & 2.8K \\
    \bottomrule
  \end{tabular}
\end{table}

\begin{table}[t]
  \centering
  \footnotesize
  \caption{Per-attempt agent effort on the operate-and-profile tasks (\S\ref{sec:eval:agent}). Each row is one independent attempt. Session Duration and Active GPU Time are in minutes.}
  \label{tab:appendix-per-attempt-operate}
  \setlength{\tabcolsep}{4pt}
  \begin{tabular}{@{}llr|rrrrr@{}}
    \toprule
    Task
      & Framework
      & Attempt
      & \makecell[br]{Session\\Duration}
      & \makecell[br]{Active\\GPU Time}
      & \makecell[br]{Agent\\Turns}
      & \makecell[br]{Per-Turn\\Context}
      & \makecell[br]{Output\\Tokens} \\
    \midrule
    \multirow{9}{*}{Getting Started}
      & \megatron    & 1st & 47.6 & 5.9 & 106 & 85.5K & 31.9K \\
      &                               & 2nd & 40.5 & 5.4 & 88 & 69.5K & 26.9K \\
      &                               & 3rd & 24.1 & 5.2 & 78 & 64.0K & 25.8K \\
      & \torchtitan  & 1st & 10.8 & 5.0 & 54 & 62.0K & 15.8K \\
      &                               & 2nd & 11.4 & 5.2 & 47 & 43.1K & 15.4K \\
      &                               & 3rd & 13.5 & 5.6 & 68 & 56.8K & 18.3K \\
      & \sys         & 1st & 7.6 & 3.9 & 20 & 33.8K & 5.4K \\
      &                               & 2nd & 6.6 & 3.0 & 26 & 37.0K & 5.8K \\
      &                               & 3rd & 5.8 & 3.1 & 28 & 37.9K & 7.6K \\
    \midrule
    \multirow{9}{*}{Train and Evaluate}
      & \megatron    & 1st & 52.3 & 29.3 & 163 & 109.2K & 45.1K \\
      &                               & 2nd & 58.8 & 36.0 & 155 & 101.4K & 53.2K \\
      &                               & 3rd & 55.5 & 38.1 & 170 & 106.4K & 52.9K \\
      & \torchtitan  & 1st & 59.5 & 36.3 & 212 & 176.6K & 75.8K \\
      &                               & 2nd & 101.6 & 87.5 & 268 & 149.8K & 97.8K \\
      &                               & 3rd & 72.5 & 28.8 & 198 & 197.1K & 109.3K \\
      & \sys         & 1st & 42.0 & 25.9 & 100 & 86.5K & 36.9K \\
      &                               & 2nd & 37.1 & 22.7 & 87 & 83.1K & 34.2K \\
      &                               & 3rd & 38.5 & 19.3 & 92 & 85.5K & 32.3K \\
    \midrule
    \multirow{9}{*}{Collect Routing Trace}
      & \megatron    & 1st & 33.3 & 8.3 & 112 & 144.3K & 102.1K \\
      &                               & 2nd & 33.3 & 5.5 & 138 & 158.8K & 102.5K \\
      &                               & 3rd & 25.1 & 5.3 & 87 & 106.4K & 74.7K \\
      & \torchtitan  & 1st & 32.8 & 10.4 & 93 & 166.0K & 84.7K \\
      &                               & 2nd & 41.4 & 11.9 & 152 & 181.7K & 113.6K \\
      &                               & 3rd & 30.8 & 9.8 & 103 & 160.3K & 74.8K \\
      & \sys         & 1st & 16.3 & 2.7 & 58 & 118.7K & 56.2K \\
      &                               & 2nd & 14.3 & 2.8 & 52 & 107.8K & 48.4K \\
      &                               & 3rd & 20.1 & 2.8 & 70 & 146.8K & 71.9K \\
    \midrule
    \multirow{9}{*}{Report Heavy Kernels}
      & \megatron    & 1st & 23.4 & 12.5 & 62 & 67.0K & 31.0K \\
      &                               & 2nd & 22.1 & 12.1 & 53 & 48.1K & 20.5K \\
      &                               & 3rd & 21.2 & 12.0 & 60 & 52.6K & 23.9K \\
      & \torchtitan  & 1st & 14.8 & 8.1 & 44 & 65.1K & 17.0K \\
      &                               & 2nd & 15.0 & 4.8 & 40 & 69.0K & 22.5K \\
      &                               & 3rd & 15.2 & 6.7 & 39 & 66.1K & 22.7K \\
      & \sys         & 1st & 11.8 & 3.7 & 42 & 50.2K & 19.0K \\
      &                               & 2nd & 10.4 & 3.6 & 36 & 44.0K & 11.6K \\
      &                               & 3rd & 12.5 & 3.5 & 43 & 49.2K & 16.0K \\
    \bottomrule
  \end{tabular}
\end{table}

\begin{table}[t]
  \centering
  \footnotesize
  \caption{Per-attempt agent effort on the new-feature tasks (\S\ref{sec:eval:agent}). Each row is one independent attempt. Session Duration and Active GPU Time are in minutes.}
  \label{tab:appendix-per-attempt-new-feature}
  \setlength{\tabcolsep}{4pt}
  \begin{tabular}{@{}llr|rrrrr@{}}
    \toprule
    Task
      & Framework
      & Attempt
      & \makecell[br]{Session\\Duration}
      & \makecell[br]{Active\\GPU Time}
      & \makecell[br]{Agent\\Turns}
      & \makecell[br]{Per-Turn\\Context}
      & \makecell[br]{Output\\Tokens} \\
    \midrule
    \multirow{9}{*}{Diff~\cite{ye2024differential}}
      & \megatron    & 1st & 47.1 & 33.7 & 62 & 118.7K & 49.4K \\
      &                               & 2nd & 56.6 & 44.9 & 125 & 165.9K & 58.3K \\
      &                               & 3rd & 46.3 & 33.6 & 141 & 103.6K & 57.1K \\
      & \torchtitan  & 1st & 49.6 & 40.3 & 61 & 96.9K & 35.4K \\
      &                               & 2nd & 57.1 & 49.2 & 53 & 105.9K & 36.0K \\
      &                               & 3rd & 48.6 & 40.1 & 58 & 103.2K & 38.7K \\
      & \sys         & 1st & 35.3 & 27.6 & 47 & 53.9K & 29.4K \\
      &                               & 2nd & 38.2 & 33.2 & 35 & 69.5K & 22.1K \\
      &                               & 3rd & 42.8 & 27.3 & 83 & 93.0K & 25.4K \\
    \midrule
    \multirow{9}{*}{DynMoE~\cite{guo2024dynamic}}
      & \megatron    & 1st & 84.2 & 45.4 & 200 & 177.9K & 125.3K \\
      &                               & 2nd & 76.7 & 49.1 & 199 & 238.1K & 115.2K \\
      &                               & 3rd & 83.8 & 62.1 & 186 & 208.0K & 94.5K \\
      & \torchtitan  & 1st & 143.6 & 97.8 & 253 & 306.1K & 161.3K \\
      &                               & 2nd & 140.6 & 94.4 & 126 & 211.4K & 178.2K \\
      &                               & 3rd & 115.5 & 90.3 & 197 & 228.8K & 113.5K \\
      & \sys         & 1st & 57.1 & 41.9 & 76 & 146.0K & 52.9K \\
      &                               & 2nd & 61.4 & 42.3 & 73 & 101.4K & 76.4K \\
      &                               & 3rd & 60.4 & 41.3 & 118 & 224.9K & 96.8K \\
    \midrule
    \multirow{9}{*}{MoBA~\cite{lu2025moba}}
      & \megatron    & 1st & 71.1 & 49.1 & 134 & 146.7K & 53.8K \\
      &                               & 2nd & 61.6 & 49.5 & 135 & 108.1K & 56.1K \\
      &                               & 3rd & 59.1 & 49.9 & 91 & 120.9K & 45.7K \\
      & \torchtitan  & 1st & 53.7 & 39.4 & 61 & 110.2K & 54.4K \\
      &                               & 2nd & 138.9 & 107.6 & 104 & 186.2K & 111.8K \\
      &                               & 3rd & 105.1 & 77.9 & 91 & 166.9K & 112.2K \\
      & \sys         & 1st & 38.0 & 27.7 & 57 & 66.4K & 32.4K \\
      &                               & 2nd & 52.9 & 32.0 & 55 & 69.0K & 22.8K \\
      &                               & 3rd & 38.7 & 26.9 & 66 & 81.5K & 50.8K \\
    \midrule
    \multirow{9}{*}{MoE++~\cite{jin2025moeplusplus}}
      & \megatron    & 1st & 144.6 & 78.0 & 829 & 516.5K & 326.4K \\
      &                               & 2nd & 88.5 & 58.7 & 145 & 170.8K & 117.0K \\
      &                               & 3rd & 75.5 & 53.6 & 140 & 188.7K & 86.9K \\
      & \torchtitan  & 1st & 95.2 & 72.1 & 85 & 134.3K & 66.0K \\
      &                               & 2nd & 71.4 & 51.9 & 87 & 164.6K & 85.3K \\
      &                               & 3rd & 63.5 & 45.2 & 125 & 173.4K & 90.4K \\
      & \sys         & 1st & 63.0 & 36.7 & 141 & 176.6K & 107.7K \\
      &                               & 2nd & 63.0 & 39.9 & 85 & 201.0K & 100.2K \\
      &                               & 3rd & 67.7 & 40.1 & 90 & 165.4K & 119.1K \\
    \bottomrule
  \end{tabular}
\end{table}

\clearpage
\par\begingroup
\subsection{Codex}\label{sec:appendix:codex}
This subsection reports per-attempt agent effort and per-task medians
for Codex across all 20 tasks.
Tables~\ref{tab:codex-qa}--\ref{tab:codex-features} report the medians,
and Tables~\ref{tab:codex-attempts-4}--\ref{tab:codex-attempts-7} give
all per-attempt values.

To test whether the results depend on the agent, we repeat the suite
with Codex (GPT-5.6 Sol at \texttt{high} reasoning effort). The framework
commits, task instructions, hardware, and three-attempt protocol are
unchanged. GPT-5.6 Sol judges new-feature correctness using the same rules
as the Claude Code evaluation (\S\ref{sec:appendix:tasks:rules}).

The overall trend holds across agents. By median Agent Turns, \sys
ranks first or ties for first on 18 of 20 Codex tasks, compared with
14 of 20 Claude Code tasks, although individual task orderings vary.
Codex also reproduces \torchtitan's MoBA OOMs in two of three attempts;
both recover and complete training.

Durations and Active GPU Time are in minutes; context and output are
in tokens. Lower is better on every metric.
\begin{table}[htbp]
  \centering
  \small
  \vspace{-7pt}
  \caption{Codex per-question agent effort on the Q\&A task suite
  (\S\ref{sec:benchmark}); Q1--Q12 are described in
  Appendix~\ref{sec:appendix-codebaseqa}. Each metric reports the
  median of three independent attempts. Lower is more efficient.
  Headers abbreviate \megatron, \torchtitan, and \sys.}
  \label{tab:codex-qa}
  \begin{tabular}{c|rrr|rrr|rrr}
    \toprule
       & \multicolumn{3}{c|}{Agent Turns}    & \multicolumn{3}{c|}{Per-Turn Context}            & \multicolumn{3}{c}{Output Tokens} \\
    \# & Megatron & Titan & Pith & Megatron & Titan & Pith & Megatron & Titan & Pith \\
    \midrule
    Q1   & 20 & 15 & \textbf{9} & 63.7K & 56.2K & \textbf{41.4K} & 10.6K & 9.6K & \textbf{6.8K} \\
    Q2   & 21 & 13 & \textbf{10} & 68.1K & 52.7K & \textbf{45.4K} & 9.2K & 6.7K & \textbf{4.9K} \\
    Q3   & \textbf{8} & 11 & \textbf{8} & 41.1K & 44.2K & \textbf{36.4K} & 4.9K & 4.0K & \textbf{3.1K} \\
    Q4   & 21 & \textbf{12} & \textbf{12} & 56.2K & 57.8K & \textbf{37.8K} & 9.7K & 8.5K & \textbf{7.2K} \\
    Q5   & 26 & 9 & \textbf{8} & 74.6K & 44.0K & \textbf{39.6K} & 9.6K & 5.3K & \textbf{4.9K} \\
    Q6   & 7 & \textbf{6} & \textbf{6} & 44.3K & 34.7K & \textbf{28.5K} & 4.3K & \textbf{3.2K} & 3.5K \\
    Q7   & 8 & 8 & \textbf{7} & 41.6K & 39.6K & \textbf{32.5K} & 3.8K & 3.5K & \textbf{3.2K} \\
    Q8   & 15 & \textbf{7} & 10 & 48.8K & 29.2K & \textbf{27.4K} & 6.5K & \textbf{3.8K} & 5.2K \\
    Q9   & 12 & 14 & \textbf{9} & 52.0K & 44.6K & \textbf{40.9K} & 5.3K & 7.3K & \textbf{5.2K} \\
    Q10  & 17 & 13 & \textbf{10} & 60.7K & 49.8K & \textbf{35.8K} & 8.5K & 7.6K & \textbf{4.4K} \\
    Q11  & 10 & \textbf{4} & 5 & 33.8K & \textbf{20.5K} & 22.8K & 5.0K & 2.8K & \textbf{2.3K} \\
    Q12  & 14 & 8 & \textbf{7} & 53.3K & 39.6K & \textbf{33.2K} & 6.3K & 4.3K & \textbf{3.7K} \\
    \bottomrule
  \end{tabular}
\end{table}
\begin{table}[htbp]
  \centering
  \small
  \caption{Codex per-task agent effort across frameworks on
  operate-and-profile tasks (\S\ref{sec:benchmark}). Each metric
  reports the median of three independent attempts. Lower is more
  efficient. Session Duration and Active GPU Time are in minutes. Per-attempt metrics in Table~\ref{tab:codex-attempts-6}.}
  \label{tab:codex-operate}
  \setlength{\tabcolsep}{3pt}
  \begin{tabular}{@{}lr|rrrrr@{}}
    \toprule
    Task
      & Framework
      & \makecell[br]{Session\\Duration}
      & \makecell[br]{Active\\GPU Time}
      & \makecell[br]{Agent\\Turns}
      & \makecell[br]{Per-Turn\\Context}
      & \makecell[br]{Output\\Tokens} \\
    \midrule
    \multirow{3}{*}{\textit{Getting Started}}
       &  \megatron    & 20.0 & 6.7 & 138 & 124.8K & 18.2K \\
       &  \torchtitan  & 8.6 & 6.6 & 45 & 55.6K & 6.1K \\
       &  \sys         & \textbf{6.0} & \textbf{3.8} & \textbf{26} & \textbf{20.4K} & \textbf{4.0K} \\
    \midrule
    \multirow{3}{*}{\textit{Train and Evaluate}}
       &  \megatron    & 65.1 & 44.5 & 317 & 166.8K & 50.9K \\
       &  \torchtitan  & 42.2 & 26.4 & 104 & 129.7K & 27.3K \\
       &  \sys         & \textbf{25.5} & \textbf{12.9} & \textbf{94} & \textbf{63.6K} & \textbf{20.6K} \\
    \midrule
    \multirow{3}{*}{\textit{Collect Routing Trace}}
       &  \megatron    & 10.1 & 6.0 & 39 & \textbf{71.0K} & \textbf{17.4K} \\
       &  \torchtitan  & 13.9 & 7.5 & 72 & 103.6K & 25.9K \\
       &  \sys         & \textbf{8.9} & \textbf{3.9} & \textbf{32} & 73.7K & 18.7K \\
    \midrule
    \multirow{3}{*}{\textit{Report Heavy Kernels}}
       &  \megatron    & 21.3 & 17.9 & 89 & 43.9K & 15.3K \\
       &  \torchtitan  & 24.3 & 17.3 & 99 & 67.6K & 20.1K \\
       &  \sys         & \textbf{11.1} & \textbf{6.8} & \textbf{42} & \textbf{43.0K} & \textbf{14.0K} \\
    \bottomrule
  \end{tabular}
\end{table}
\begin{table}[htbp]
  \centering
  \small
  \vspace{-5pt}
  \caption{Codex per-task agent effort across frameworks on new-feature
  tasks (\S\ref{sec:benchmark}). Each metric reports the median of
  three independent attempts. Lower is more efficient. Session Duration
  and Active GPU Time are in minutes. Per-attempt metrics in
  Table~\ref{tab:codex-attempts-7}.
  }
  \label{tab:codex-features}
  \setlength{\tabcolsep}{3pt}
  \begin{tabular}{@{}lr|rrrrr@{}}
    \toprule
    Task
      & Framework
      & \makecell[br]{Session\\Duration}
      & \makecell[br]{Active\\GPU Time}
      & \makecell[br]{Agent\\Turns}
      & \makecell[br]{Per-Turn\\Context}
      & \makecell[br]{Output\\Tokens} \\
    \midrule
    \multirow{3}{*}{\makecell[l]{\textit{Differential Transformer}\\(Diff)~\cite{ye2024differential}}}
       &  \megatron    & 49.8 & 44.7 & 163 & 136.0K & 27.0K \\
       &  \torchtitan  & 50.7 & 46.6 & 119 & 96.4K & 17.4K \\
       &  \sys         & \textbf{36.3} & \textbf{33.8} & \textbf{112} & \textbf{78.5K} & \textbf{14.6K} \\
    \midrule
    \multirow{3}{*}{\makecell[l]{\textit{Dynamic Mixture of Experts}\\(DynMoE)~\cite{guo2024dynamic}}}
       &  \megatron    & 66.9 & 56.6 & 203 & 152.5K & 41.4K \\
       &  \torchtitan  & 105.4 & 93.8 & 257 & 206.3K & 51.4K \\
       &  \sys         & \textbf{46.9} & \textbf{39.5} & \textbf{148} & \textbf{141.8K} & \textbf{32.1K} \\
    \midrule
    \multirow{3}{*}{\makecell[l]{\textit{Mixture of Block Attention}\\(MoBA)~\cite{lu2025moba}}}
       &  \megatron    & 78.8 & 70.1 & 200 & 189.4K & 51.3K \\
       &  \torchtitan  & 74.7 & 67.7 & 173 & 134.6K & 38.1K \\
       &  \sys         & \textbf{47.9} & \textbf{42.8} & \textbf{95} & \textbf{102.8K} & \textbf{26.9K} \\
    \midrule
    \multirow{3}{*}{\textit{MoE++}~\cite{jin2025moeplusplus}}
       &  \megatron    & 141.0 & 123.1 & 527 & 168.6K & 74.6K \\
       &  \torchtitan  & 76.2 & 66.4 & 226 & 173.4K & 44.8K \\
       &  \sys         & \textbf{43.9} & \textbf{37.5} & \textbf{200} & \textbf{165.4K} & \textbf{41.8K} \\
    \bottomrule
  \end{tabular}
\end{table}
\clearpage
\begin{table}[t]
  \centering
  \footnotesize
  \caption{Codex per-attempt agent effort on the Q\&A tasks Q1--Q6 (\S\ref{sec:appendix:codex}). Each row is one independent attempt. Session Duration is in minutes.}
  \label{tab:codex-attempts-4}
  \setlength{\tabcolsep}{4pt}
  \begin{tabular}{@{}llr|rrrr@{}}
    \toprule
    Task
      & Framework
      & Attempt
      & \makecell[br]{Session\\Duration}
      & \makecell[br]{Agent\\Turns}
      & \makecell[br]{Per-Turn\\Context}
      & \makecell[br]{Output\\Tokens} \\
    \midrule
    \multirow{9}{*}{Q1: Process Groups / Device Mesh}
       &  \megatron      &  1st  & 4.8 & 20 & 68.9K & 13.2K \\
       &                                 &  2nd  & 3.7 & 20 & 63.1K & 10.6K \\
       &                                 &  3rd  & 3.4 & 18 & 63.7K & 8.6K \\
       &  \torchtitan    &  1st  & 3.5 & 17 & 56.2K & 9.6K \\
       &                                 &  2nd  & 2.4 & 13 & 54.5K & 6.8K \\
       &                                 &  3rd  & 3.9 & 15 & 65.2K & 10.0K \\
       &  \sys           &  1st  & 2.4 & 9 & 42.4K & 6.8K \\
       &                                 &  2nd  & 2.0 & 7 & 33.0K & 5.9K \\
       &                                 &  3rd  & 2.9 & 14 & 41.4K & 7.9K \\
    \midrule
    \multirow{9}{*}{Q2: Configuration Propagation}
       &  \megatron      &  1st  & 3.5 & 20 & 74.3K & 9.2K \\
       &                                 &  2nd  & 4.7 & 27 & 68.1K & 10.3K \\
       &                                 &  3rd  & 3.8 & 21 & 60.0K & 8.3K \\
       &  \torchtitan    &  1st  & 2.6 & 13 & 49.3K & 6.5K \\
       &                                 &  2nd  & 2.4 & 12 & 52.7K & 6.7K \\
       &                                 &  3rd  & 2.5 & 14 & 55.3K & 6.9K \\
       &  \sys           &  1st  & 1.5 & 9 & 42.6K & 4.2K \\
       &                                 &  2nd  & 1.8 & 11 & 48.5K & 5.0K \\
       &                                 &  3rd  & 1.9 & 10 & 45.4K & 4.9K \\
    \midrule
    \multirow{9}{*}{Q3: Data Loading \& Sharding}
       &  \megatron      &  1st  & 1.0 & 7 & 41.1K & 2.7K \\
       &                                 &  2nd  & 1.9 & 10 & 44.4K & 4.9K \\
       &                                 &  3rd  & 1.8 & 8 & 37.9K & 5.3K \\
       &  \torchtitan    &  1st  & 1.6 & 11 & 39.7K & 4.0K \\
       &                                 &  2nd  & 2.3 & 18 & 49.6K & 5.5K \\
       &                                 &  3rd  & 1.5 & 11 & 44.2K & 3.9K \\
       &  \sys           &  1st  & 1.3 & 8 & 36.4K & 3.1K \\
       &                                 &  2nd  & 1.7 & 10 & 40.1K & 4.5K \\
       &                                 &  3rd  & 1.2 & 7 & 35.8K & 3.1K \\
    \midrule
    \multirow{9}{*}{Q4: Distributed Seed Management}
       &  \megatron      &  1st  & 3.7 & 21 & 47.6K & 9.7K \\
       &                                 &  2nd  & 3.1 & 17 & 56.2K & 7.9K \\
       &                                 &  3rd  & 5.1 & 24 & 71.2K & 12.6K \\
       &  \torchtitan    &  1st  & 2.5 & 11 & 49.3K & 6.5K \\
       &                                 &  2nd  & 3.4 & 15 & 57.8K & 9.9K \\
       &                                 &  3rd  & 3.2 & 12 & 62.8K & 8.5K \\
       &  \sys           &  1st  & 2.7 & 12 & 31.5K & 7.2K \\
       &                                 &  2nd  & 2.6 & 12 & 50.7K & 7.6K \\
       &                                 &  3rd  & 2.3 & 7 & 37.8K & 7.1K \\
    \midrule
    \multirow{9}{*}{Q5: Attention Kernel Dispatch}
       &  \megatron      &  1st  & 4.7 & 35 & 79.1K & 11.2K \\
       &                                 &  2nd  & 2.8 & 15 & 61.1K & 7.3K \\
       &                                 &  3rd  & 3.8 & 26 & 74.6K & 9.6K \\
       &  \torchtitan    &  1st  & 1.8 & 9 & 40.4K & 5.1K \\
       &                                 &  2nd  & 1.9 & 9 & 49.3K & 5.3K \\
       &                                 &  3rd  & 1.9 & 9 & 44.0K & 5.6K \\
       &  \sys           &  1st  & 1.6 & 8 & 39.0K & 4.2K \\
       &                                 &  2nd  & 2.1 & 10 & 47.6K & 5.5K \\
       &                                 &  3rd  & 1.7 & 8 & 39.6K & 4.9K \\
    \midrule
    \multirow{9}{*}{Q6: RoPE Implementation}
       &  \megatron      &  1st  & 1.6 & 9 & 45.2K & 4.3K \\
       &                                 &  2nd  & 1.4 & 6 & 44.3K & 4.0K \\
       &                                 &  3rd  & 1.6 & 7 & 35.6K & 4.5K \\
       &  \torchtitan    &  1st  & 1.1 & 6 & 34.7K & 3.0K \\
       &                                 &  2nd  & 1.2 & 7 & 35.9K & 3.4K \\
       &                                 &  3rd  & 1.1 & 5 & 31.7K & 3.2K \\
       &  \sys           &  1st  & 2.1 & 6 & 36.7K & 6.3K \\
       &                                 &  2nd  & 1.4 & 7 & 28.5K & 3.5K \\
       &                                 &  3rd  & 1.1 & 5 & 26.8K & 3.3K \\
    \bottomrule
  \end{tabular}
\end{table}
\begin{table}[t]
  \centering
  \footnotesize
  \caption{Codex per-attempt agent effort on the Q\&A tasks (continued, Q7--Q12).}
  \label{tab:codex-attempts-5}
  \setlength{\tabcolsep}{4pt}
  \begin{tabular}{@{}llr|rrrr@{}}
    \toprule
    Task
      & Framework
      & Attempt
      & \makecell[br]{Session\\Duration}
      & \makecell[br]{Agent\\Turns}
      & \makecell[br]{Per-Turn\\Context}
      & \makecell[br]{Output\\Tokens} \\
    \midrule
    \multirow{9}{*}{Q7: SwiGLU / MLP Block}
       &  \megatron      &  1st  & 1.5 & 8 & 41.6K & 4.0K \\
       &                                 &  2nd  & 1.4 & 8 & 39.6K & 3.8K \\
       &                                 &  3rd  & 1.3 & 7 & 41.8K & 3.4K \\
       &  \torchtitan    &  1st  & 1.3 & 8 & 40.3K & 3.5K \\
       &                                 &  2nd  & 1.4 & 8 & 35.4K & 3.9K \\
       &                                 &  3rd  & 1.4 & 6 & 39.6K & 3.5K \\
       &  \sys           &  1st  & 1.6 & 9 & 38.7K & 4.4K \\
       &                                 &  2nd  & 1.5 & 6 & 32.5K & 3.2K \\
       &                                 &  3rd  & 1.3 & 7 & 29.7K & 3.1K \\
    \midrule
    \multirow{9}{*}{Q8: Normalization Placement}
       &  \megatron      &  1st  & 2.3 & 15 & 45.5K & 6.3K \\
       &                                 &  2nd  & 2.5 & 15 & 48.8K & 6.5K \\
       &                                 &  3rd  & 3.0 & 14 & 54.5K & 7.8K \\
       &  \torchtitan    &  1st  & 2.4 & 13 & 35.7K & 5.4K \\
       &                                 &  2nd  & 1.1 & 5 & 29.2K & 3.1K \\
       &                                 &  3rd  & 1.5 & 7 & 28.6K & 3.8K \\
       &  \sys           &  1st  & 3.6 & 14 & 31.2K & 5.7K \\
       &                                 &  2nd  & 1.9 & 10 & 25.8K & 5.2K \\
       &                                 &  3rd  & 1.5 & 10 & 27.4K & 4.2K \\
    \midrule
    \multirow{9}{*}{Q9: Context / Sequence Parallelism}
       &  \megatron      &  1st  & 1.8 & 12 & 52.0K & 4.9K \\
       &                                 &  2nd  & 2.9 & 16 & 58.8K & 8.1K \\
       &                                 &  3rd  & 1.9 & 11 & 44.8K & 5.3K \\
       &  \torchtitan    &  1st  & 2.9 & 14 & 44.3K & 7.4K \\
       &                                 &  2nd  & 3.1 & 14 & 46.4K & 7.2K \\
       &                                 &  3rd  & 2.9 & 13 & 44.6K & 7.3K \\
       &  \sys           &  1st  & 2.2 & 9 & 39.6K & 5.2K \\
       &                                 &  2nd  & 2.9 & 13 & 44.1K & 7.7K \\
       &                                 &  3rd  & 1.8 & 8 & 40.9K & 5.0K \\
    \midrule
    \multirow{9}{*}{Q10: FSDP / DDP Wrapping}
       &  \megatron      &  1st  & 3.2 & 14 & 60.7K & 8.7K \\
       &                                 &  2nd  & 3.8 & 17 & 63.7K & 8.5K \\
       &                                 &  3rd  & 2.9 & 18 & 57.6K & 7.0K \\
       &  \torchtitan    &  1st  & 2.8 & 14 & 49.8K & 7.8K \\
       &                                 &  2nd  & 2.3 & 12 & 58.0K & 6.3K \\
       &                                 &  3rd  & 2.8 & 13 & 38.3K & 7.6K \\
       &  \sys           &  1st  & 1.7 & 10 & 38.4K & 4.4K \\
       &                                 &  2nd  & 1.4 & 6 & 28.2K & 4.1K \\
       &                                 &  3rd  & 2.8 & 15 & 35.8K & 7.9K \\
    \midrule
    \multirow{9}{*}{Q11: Global Gradient Clipping}
       &  \megatron      &  1st  & 1.4 & 7 & 28.9K & 3.8K \\
       &                                 &  2nd  & 1.8 & 10 & 33.8K & 5.0K \\
       &                                 &  3rd  & 2.0 & 10 & 34.6K & 5.6K \\
       &  \torchtitan    &  1st  & 0.8 & 4 & 20.5K & 2.2K \\
       &                                 &  2nd  & 1.3 & 5 & 19.2K & 3.7K \\
       &                                 &  3rd  & 1.1 & 4 & 23.6K & 2.8K \\
       &  \sys           &  1st  & 0.9 & 5 & 23.0K & 2.2K \\
       &                                 &  2nd  & 0.9 & 4 & 19.0K & 2.3K \\
       &                                 &  3rd  & 1.2 & 6 & 22.8K & 3.4K \\
    \midrule
    \multirow{9}{*}{Q12: Distributed Checkpoint Serialization}
       &  \megatron      &  1st  & 2.6 & 18 & 53.3K & 6.9K \\
       &                                 &  2nd  & 2.1 & 14 & 50.2K & 5.4K \\
       &                                 &  3rd  & 2.5 & 14 & 60.8K & 6.3K \\
       &  \torchtitan    &  1st  & 1.3 & 7 & 38.6K & 3.6K \\
       &                                 &  2nd  & 1.6 & 8 & 39.6K & 4.3K \\
       &                                 &  3rd  & 2.4 & 11 & 49.3K & 6.7K \\
       &  \sys           &  1st  & 1.4 & 7 & 33.2K & 3.7K \\
       &                                 &  2nd  & 1.3 & 6 & 35.4K & 3.1K \\
       &                                 &  3rd  & 2.5 & 9 & 32.7K & 5.7K \\
    \bottomrule
  \end{tabular}
\end{table}
\begin{table}[t]
  \centering
  \footnotesize
  \caption{Codex per-attempt agent effort on the operate-and-profile tasks (\S\ref{sec:appendix:codex}). Each row is one independent attempt. Session Duration and Active GPU Time are in minutes.}
  \label{tab:codex-attempts-6}
  \setlength{\tabcolsep}{4pt}
  \begin{tabular}{@{}llr|rrrrr@{}}
    \toprule
    Task
      & Framework
      & Attempt
      & \makecell[br]{Session\\Duration}
      & \makecell[br]{Active\\GPU Time}
      & \makecell[br]{Agent\\Turns}
      & \makecell[br]{Per-Turn\\Context}
      & \makecell[br]{Output\\Tokens} \\
    \midrule
    \multirow{9}{*}{Getting Started}
       &  \megatron     &  1st  & 15.2 & 6.7 & 69 & 79.8K & 10.8K \\
       &                                 &  2nd  & 33.2 & 8.0 & 284 & 127.6K & 27.6K \\
       &                                 &  3rd  & 20.0 & 6.7 & 138 & 124.8K & 18.2K \\
       &  \torchtitan   &  1st  & 8.6 & 6.6 & 36 & 49.5K & 6.1K \\
       &                                 &  2nd  & 8.7 & 6.5 & 45 & 61.7K & 8.7K \\
       &                                 &  3rd  & 8.2 & 6.6 & 45 & 55.6K & 5.9K \\
       &  \sys          &  1st  & 6.0 & 3.8 & 28 & 20.4K & 4.0K \\
       &                                 &  2nd  & 6.0 & 3.8 & 26 & 35.2K & 4.7K \\
       &                                 &  3rd  & 4.2 & 3.3 & 21 & 18.7K & 3.6K \\
    \midrule
    \multirow{9}{*}{Train and Evaluate}
       &  \megatron     &  1st  & 69.6 & 44.5 & 317 & 190.9K & 50.9K \\
       &                                 &  2nd  & 65.1 & 49.4 & 221 & 113.7K & 34.3K \\
       &                                 &  3rd  & 57.1 & 35.5 & 321 & 166.8K & 53.6K \\
       &  \torchtitan   &  1st  & 43.8 & 31.2 & 104 & 142.0K & 27.3K \\
       &                                 &  2nd  & 42.2 & 26.4 & 146 & 129.7K & 36.4K \\
       &                                 &  3rd  & 27.2 & 15.8 & 103 & 121.3K & 26.0K \\
       &  \sys          &  1st  & 26.7 & 12.7 & 108 & 61.4K & 22.7K \\
       &                                 &  2nd  & 25.5 & 12.9 & 94 & 63.6K & 19.6K \\
       &                                 &  3rd  & 25.0 & 13.1 & 72 & 71.0K & 20.6K \\
    \midrule
    \multirow{9}{*}{Collect Routing Trace}
       &  \megatron     &  1st  & 10.1 & 6.0 & 47 & 79.9K & 17.4K \\
       &                                 &  2nd  & 10.2 & 6.0 & 39 & 71.0K & 18.5K \\
       &                                 &  3rd  & 9.5 & 5.2 & 33 & 69.3K & 16.9K \\
       &  \torchtitan   &  1st  & 12.7 & 7.5 & 62 & 106.4K & 20.6K \\
       &                                 &  2nd  & 26.1 & 20.5 & 113 & 98.9K & 25.9K \\
       &                                 &  3rd  & 13.9 & 7.4 & 72 & 103.6K & 26.6K \\
       &  \sys          &  1st  & 7.8 & 3.3 & 30 & 75.5K & 18.7K \\
       &                                 &  2nd  & 12.3 & 3.9 & 32 & 71.2K & 21.4K \\
       &                                 &  3rd  & 8.9 & 3.9 & 35 & 73.7K & 17.0K \\
    \midrule
    \multirow{9}{*}{Report Heavy Kernels}
       &  \megatron     &  1st  & 21.3 & 18.8 & 50 & 43.9K & 13.9K \\
       &                                 &  2nd  & 20.9 & 17.9 & 93 & 44.9K & 15.3K \\
       &                                 &  3rd  & 22.1 & 14.4 & 89 & 42.6K & 17.6K \\
       &  \torchtitan   &  1st  & 24.3 & 17.3 & 99 & 68.8K & 20.1K \\
       &                                 &  2nd  & 21.1 & 15.1 & 62 & 67.6K & 21.1K \\
       &                                 &  3rd  & 24.5 & 18.6 & 105 & 52.3K & 18.3K \\
       &  \sys          &  1st  & 11.9 & 6.6 & 41 & 43.0K & 13.3K \\
       &                                 &  2nd  & 11.1 & 6.8 & 51 & 43.7K & 16.2K \\
       &                                 &  3rd  & 9.4 & 6.8 & 42 & 37.1K & 14.0K \\
    \bottomrule
  \end{tabular}
\end{table}
\begin{table}[t]
  \centering
  \footnotesize
  \caption{Codex per-attempt agent effort on the new-feature tasks (\S\ref{sec:appendix:codex}). Each row is one independent attempt. Session Duration and Active GPU Time are in minutes.}
  \label{tab:codex-attempts-7}
  \setlength{\tabcolsep}{4pt}
  \begin{tabular}{@{}llr|rrrrr@{}}
    \toprule
    Task
      & Framework
      & Attempt
      & \makecell[br]{Session\\Duration}
      & \makecell[br]{Active\\GPU Time}
      & \makecell[br]{Agent\\Turns}
      & \makecell[br]{Per-Turn\\Context}
      & \makecell[br]{Output\\Tokens} \\
    \midrule
    \multirow{9}{*}{Diff~\cite{ye2024differential}}
       &  \megatron     &  1st  & 54.8 & 48.9 & 163 & 128.4K & 27.0K \\
       &                                 &  2nd  & 49.2 & 44.4 & 117 & 137.0K & 22.3K \\
       &                                 &  3rd  & 49.8 & 44.7 & 164 & 136.0K & 27.8K \\
       &  \torchtitan   &  1st  & 49.9 & 46.9 & 119 & 86.5K & 15.3K \\
       &                                 &  2nd  & 50.7 & 46.6 & 119 & 97.3K & 17.4K \\
       &                                 &  3rd  & 51.0 & 46.6 & 113 & 96.4K & 19.4K \\
       &  \sys          &  1st  & 35.2 & 33.1 & 143 & 68.0K & 14.6K \\
       &                                 &  2nd  & 36.3 & 33.8 & 91 & 86.7K & 12.8K \\
       &                                 &  3rd  & 36.4 & 33.8 & 112 & 78.5K & 14.9K \\
    \midrule
    \multirow{9}{*}{DynMoE~\cite{guo2024dynamic}}
       &  \megatron     &  1st  & 59.1 & 50.3 & 170 & 152.5K & 40.3K \\
       &                                 &  2nd  & 85.6 & 76.9 & 215 & 171.7K & 41.4K \\
       &                                 &  3rd  & 66.9 & 56.6 & 203 & 60.8K & 43.8K \\
       &  \torchtitan   &  1st  & 78.5 & 67.2 & 257 & 206.3K & 47.3K \\
       &                                 &  2nd  & 109.6 & 98.1 & 208 & 202.6K & 51.4K \\
       &                                 &  3rd  & 105.4 & 93.8 & 310 & 210.0K & 53.0K \\
       &  \sys          &  1st  & 46.9 & 39.7 & 148 & 162.5K & 32.1K \\
       &                                 &  2nd  & 46.9 & 39.5 & 98 & 141.8K & 31.1K \\
       &                                 &  3rd  & 47.0 & 38.2 & 186 & 116.6K & 37.6K \\
    \midrule
    \multirow{9}{*}{MoBA~\cite{lu2025moba}}
       &  \megatron     &  1st  & 78.8 & 70.1 & 163 & 165.6K & 51.3K \\
       &                                 &  2nd  & 62.7 & 53.6 & 200 & 191.4K & 37.9K \\
       &                                 &  3rd  & 103.7 & 90.8 & 223 & 189.4K & 58.5K \\
       &  \torchtitan   &  1st  & 56.2 & 47.4 & 152 & 134.6K & 38.1K \\
       &                                 &  2nd  & 74.7 & 67.7 & 195 & 139.8K & 31.8K \\
       &                                 &  3rd  & 87.8 & 78.4 & 173 & 125.2K & 46.0K \\
       &  \sys          &  1st  & 47.4 & 42.8 & 82 & 86.4K & 24.7K \\
       &                                 &  2nd  & 47.9 & 42.7 & 119 & 109.3K & 29.0K \\
       &                                 &  3rd  & 48.3 & 43.3 & 95 & 102.8K & 26.9K \\
    \midrule
    \multirow{9}{*}{MoE++~\cite{jin2025moeplusplus}}
       &  \megatron     &  1st  & 141.0 & 123.1 & 527 & 168.6K & 74.6K \\
       &                                 &  2nd  & 71.2 & 62.5 & 203 & 182.4K & 38.3K \\
       &                                 &  3rd  & 161.7 & 133.6 & 775 & 146.4K & 128.7K \\
       &  \torchtitan   &  1st  & 68.6 & 61.2 & 281 & 168.3K & 42.8K \\
       &                                 &  2nd  & 76.2 & 66.4 & 226 & 173.4K & 45.4K \\
       &                                 &  3rd  & 82.4 & 72.7 & 225 & 197.3K & 44.8K \\
       &  \sys          &  1st  & 43.0 & 37.5 & 200 & 165.4K & 41.8K \\
       &                                 &  2nd  & 43.9 & 35.8 & 125 & 151.2K & 30.8K \\
       &                                 &  3rd  & 64.5 & 48.0 & 233 & 214.6K & 50.0K \\
    \bottomrule
  \end{tabular}
\end{table}
\clearpage
\subsection{Run-to-Run Variation}\label{sec:appendix:dispersion}
We report per-task medians to reduce sensitivity to extreme attempts.
For example, \megatron's three MoE++ attempts use 140, 145, and 829
Agent Turns, compared with 85, 90, and 141 on \sys.
Tables~\ref{tab:ranges-claude} and~\ref{tab:ranges-codex} report Agent
Turns as median [minimum--maximum] over the three independent attempts.
Tables~\ref{tab:gpu-ranges-claude} and~\ref{tab:gpu-ranges-codex} give
the same summary for new-feature Active GPU Time. On all four new-feature tasks,
every \sys attempt uses less Active GPU Time than every attempt on
either baseline, under both agents. Agent-Turn ranges overlap on some
tasks, illustrating the variation behind the median comparisons.
{\small\setlength{\tabcolsep}{4pt}
\begin{longtable}{lrrr}
\caption{Claude Agent Turns: median [minimum--maximum] across three attempts.}\label{tab:ranges-claude}\\
\toprule
Task & \megatron & \torchtitan & \sys \\
\midrule\endfirsthead
\toprule
Task & \megatron & \torchtitan & \sys \\
\midrule\endhead
\bottomrule\endfoot
Q1 & 33 [26--34] & 18 [15--19] & 15 [13--17] \\
Q2 & 54 [37--54] & 28 [25--34] & 18 [18--28] \\
Q3 & 14 [10--14] & 14 [14--21] & 13 [8--15] \\
Q4 & 36 [34--37] & 26 [22--32] & 16 [15--19] \\
Q5 & 48 [27--63] & 19 [15--20] & 21 [20--24] \\
Q6 & 12 [12--12] & 4 [4--4] & 11 [8--18] \\
Q7 & 9 [8--10] & 12 [11--13] & 14 [13--15] \\
Q8 & 20 [15--25] & 11 [8--14] & 9 [9--11] \\
Q9 & 34 [29--60] & 16 [16--20] & 16 [16--22] \\
Q10 & 19 [16--21] & 17 [16--24] & 11 [9--12] \\
Q11 & 10 [10--12] & 8 [7--9] & 6 [6--7] \\
Q12 & 18 [17--19] & 5 [5--5] & 10 [9--10] \\
Start & 88 [78--106] & 54 [47--68] & 26 [20--28] \\
Train/eval & 163 [155--170] & 212 [198--268] & 92 [87--100] \\
Routing & 112 [87--138] & 103 [93--152] & 58 [52--70] \\
Kernels & 60 [53--62] & 40 [39--44] & 42 [36--43] \\
Diff & 125 [62--141] & 58 [53--61] & 47 [35--83] \\
DynMoE & 199 [186--200] & 197 [126--253] & 76 [73--118] \\
MoBA & 134 [91--135] & 91 [61--104] & 57 [55--66] \\
MoE++ & 145 [140--829] & 87 [85--125] & 90 [85--141] \\
\end{longtable}
}
{\small\setlength{\tabcolsep}{4pt}
\begin{longtable}{lrrr}
\caption{Codex Agent Turns: median [minimum--maximum] across three attempts.}\label{tab:ranges-codex}\\
\toprule
Task & \megatron & \torchtitan & \sys \\
\midrule\endfirsthead
\toprule
Task & \megatron & \torchtitan & \sys \\
\midrule\endhead
\bottomrule\endfoot
Q1 & 20 [18--20] & 15 [13--17] & 9 [7--14] \\
Q2 & 21 [20--27] & 13 [12--14] & 10 [9--11] \\
Q3 & 8 [7--10] & 11 [11--18] & 8 [7--10] \\
Q4 & 21 [17--24] & 12 [11--15] & 12 [7--12] \\
Q5 & 26 [15--35] & 9 [9--9] & 8 [8--10] \\
Q6 & 7 [6--9] & 6 [5--7] & 6 [5--7] \\
Q7 & 8 [7--8] & 8 [6--8] & 7 [6--9] \\
Q8 & 15 [14--15] & 7 [5--13] & 10 [10--14] \\
Q9 & 12 [11--16] & 14 [13--14] & 9 [8--13] \\
Q10 & 17 [14--18] & 13 [12--14] & 10 [6--15] \\
Q11 & 10 [7--10] & 4 [4--5] & 5 [4--6] \\
Q12 & 14 [14--18] & 8 [7--11] & 7 [6--9] \\
Start & 138 [69--284] & 45 [36--45] & 26 [21--28] \\
Train/eval & 317 [221--321] & 104 [103--146] & 94 [72--108] \\
Routing & 39 [33--47] & 72 [62--113] & 32 [30--35] \\
Kernels & 89 [50--93] & 99 [62--105] & 42 [41--51] \\
Diff & 163 [117--164] & 119 [113--119] & 112 [91--143] \\
DynMoE & 203 [170--215] & 257 [208--310] & 148 [98--186] \\
MoBA & 200 [163--223] & 173 [152--195] & 95 [82--119] \\
MoE++ & 527 [203--775] & 226 [225--281] & 200 [125--233] \\
\end{longtable}
}
{\small\setlength{\tabcolsep}{4pt}
\begin{longtable}{lrrr}
\caption{Claude new-feature Active GPU Time: median [minimum--maximum] in minutes.}\label{tab:gpu-ranges-claude}\\
\toprule
Task & \megatron & \torchtitan & \sys \\
\midrule\endfirsthead
\toprule
Task & \megatron & \torchtitan & \sys \\
\midrule\endhead
\bottomrule\endfoot
Diff & 33.7 [33.6--44.9] & 40.3 [40.1--49.2] & 27.6 [27.3--33.2] \\
DynMoE & 49.1 [45.4--62.1] & 94.4 [90.3--97.8] & 41.9 [41.3--42.3] \\
MoBA & 49.5 [49.1--49.9] & 77.9 [39.4--107.6] & 27.7 [26.9--32.0] \\
MoE++ & 58.7 [53.6--78.0] & 51.9 [45.2--72.1] & 39.9 [36.7--40.1] \\
\end{longtable}
}
{\small\setlength{\tabcolsep}{4pt}
\begin{longtable}{lrrr}
\caption{Codex new-feature Active GPU Time: median [minimum--maximum] in minutes.}\label{tab:gpu-ranges-codex}\\
\toprule
Task & \megatron & \torchtitan & \sys \\
\midrule\endfirsthead
\toprule
Task & \megatron & \torchtitan & \sys \\
\midrule\endhead
\bottomrule\endfoot
Diff & 44.7 [44.4--48.9] & 46.6 [46.6--46.9] & 33.8 [33.1--33.8] \\
DynMoE & 56.6 [50.3--76.9] & 93.8 [67.2--98.1] & 39.5 [38.2--39.7] \\
MoBA & 70.1 [53.6--90.8] & 67.7 [47.4--78.4] & 42.8 [42.7--43.3] \\
MoE++ & 123.1 [62.5--133.6] & 66.4 [61.2--72.7] & 37.5 [35.8--48.0] \\
\end{longtable}
}

\par\endgroup

\clearpage
\par\begingroup
\section{Agent Skills on Production Frameworks}\label{sec:appendix:ablations}
\label{sec:appendix:baseline-skills}
We tailor \texttt{capture-nsys-profile} to \megatron and \torchtitan,
then repeat Report Heavy Kernels with and without the skill. Each
condition uses Codex for three attempts with a fixed harness, following
the skill-removal protocol in \S\ref{sec:eval:skills-ablation}.
Table~\ref{tab:baseline-skills} reports the medians.

On \torchtitan, the skill reduces Agent Turns from 99 to 53 and Active
GPU Time from 17.3 to 9.1 minutes. Without the skill, the agent builds
step-scoped profiling hooks in all three attempts, each with a different
design, and runs training repeatedly to get them right. The skill
provides this procedure directly. On \megatron, Agent Turns increase
from 89 to 98 while GPU time stays near parity (17.9 versus 17.6 minutes).
Its existing profiling flags already expose the procedure, so reading
the skill adds overhead. Skills help when procedural knowledge is
missing; their effect depends on what the framework already exposes.

{\small\setlength{\tabcolsep}{4pt}
\begin{longtable}{llrrrrr}
\caption{Baseline profiling-skills ablation: medians of three Codex attempts. Durations are in minutes, and context and output are in tokens.}\label{tab:baseline-skills}\\
\toprule
Framework & Skill & Duration & GPU time & Turns & Context & Output \\
\midrule\endfirsthead
\toprule
Framework & Skill & Duration & GPU time & Turns & Context & Output \\
\midrule\endhead
\bottomrule\endfoot
\megatron & off & 21.3 & 17.9 & 89 & 43.9K & 15.3K \\
 & on & 23.5 & 17.6 & 98 & 37.9K & 16.1K \\
\torchtitan & off & 24.3 & 17.3 & 99 & 67.6K & 20.1K \\
 & on & 12.7 & 9.1 & 53 & 39.0K & 14.0K \\
\end{longtable}
}
\par\endgroup

\par\begingroup
\section{Model FLOPs Utilization}
\label{sec:appendix:mfu}

We compute MFU~\cite{chowdhery2023palm} as
$T \cdot F_{\mathrm{token}} / (N_{\mathrm{GPU}} \cdot P_{\mathrm{peak}})$,
where $T$ is the aggregate throughput in tokens per second and
$P_{\mathrm{peak}} = 989.4$~TFLOPS is the dense BF16 peak of an H200.
$F_{\mathrm{token}}$ follows the \texttt{num\_floating\_point\_operations}
formula of \megatron \texttt{core\_v0.20.0} with three corrections:
gated experts count three weight matrices (gate, up, and down), the
router GEMM is included, and sliding-window layers attend over their
window rather than the full causal
context\footnote{\url{https://github.com/NVIDIA/Megatron-LM/issues/1725}}.
This gives 23.08 and 56.91~GFLOP per token for Qwen3-30B-A3B at 4K and
32K, 32.71~GFLOP per token for GPT-OSS-120B at 4K, and 16.41~GFLOP per
token for DeepSeek-V2-Lite at 4K. Without the
corrections, the formula undercounts \mbox{GPT-OSS-120B} by about 17\%; for
Qwen3-30B-A3B, only the router term changes.
\par\endgroup

\end{document}